\documentclass[runningheads]{llncs}
\usepackage{graphicx,array}
\usepackage{amsmath,amssymb} 
\usepackage{color}
\usepackage{array,multirow}
\usepackage[width=122mm,left=12mm,paperwidth=146mm,height=193mm,top=12mm,paperheight=217mm]{geometry}
\begin{document}

\newcommand\TODO[1]{\textcolor{red}{[[#1]]}}

\pagestyle{headings}
\mainmatter

\title{Laplacian Pyramid Reconstruction and Refinement for Semantic Segmentation} 

\author{Golnaz Ghiasi and Charless C. Fowlkes\\ {\tt \{gghiasi,fowlkes\}@ics.uci.edu}}
\institute{Dept. of Computer Science,\\ University of California, Irvine}
\titlerunning{Laplacian reconstruction and refinement}
\authorrunning{G. Ghiasi and C. Fowlkes}

\maketitle

\begin{abstract}
CNN architectures have terrific recognition performance but rely on spatial
pooling which makes it difficult to adapt them to tasks that require dense,
pixel-accurate labeling. This paper makes two contributions: (1) We demonstrate
that while the apparent spatial resolution of convolutional feature maps is
low, the high-dimensional feature representation contains significant sub-pixel
localization information. (2) We describe a multi-resolution reconstruction
architecture based on a Laplacian pyramid that uses skip connections from
higher resolution feature maps and multiplicative gating to successively refine
segment boundaries reconstructed from lower-resolution maps.  This approach
yields state-of-the-art semantic segmentation results on the PASCAL VOC and
Cityscapes segmentation benchmarks without resorting to more complex
random-field inference or instance detection driven architectures.
\end{abstract}

\section{Introduction}

\begin{figure}[ht]
\begin{center}
	\includegraphics[ width=0.95\linewidth, keepaspectratio,]{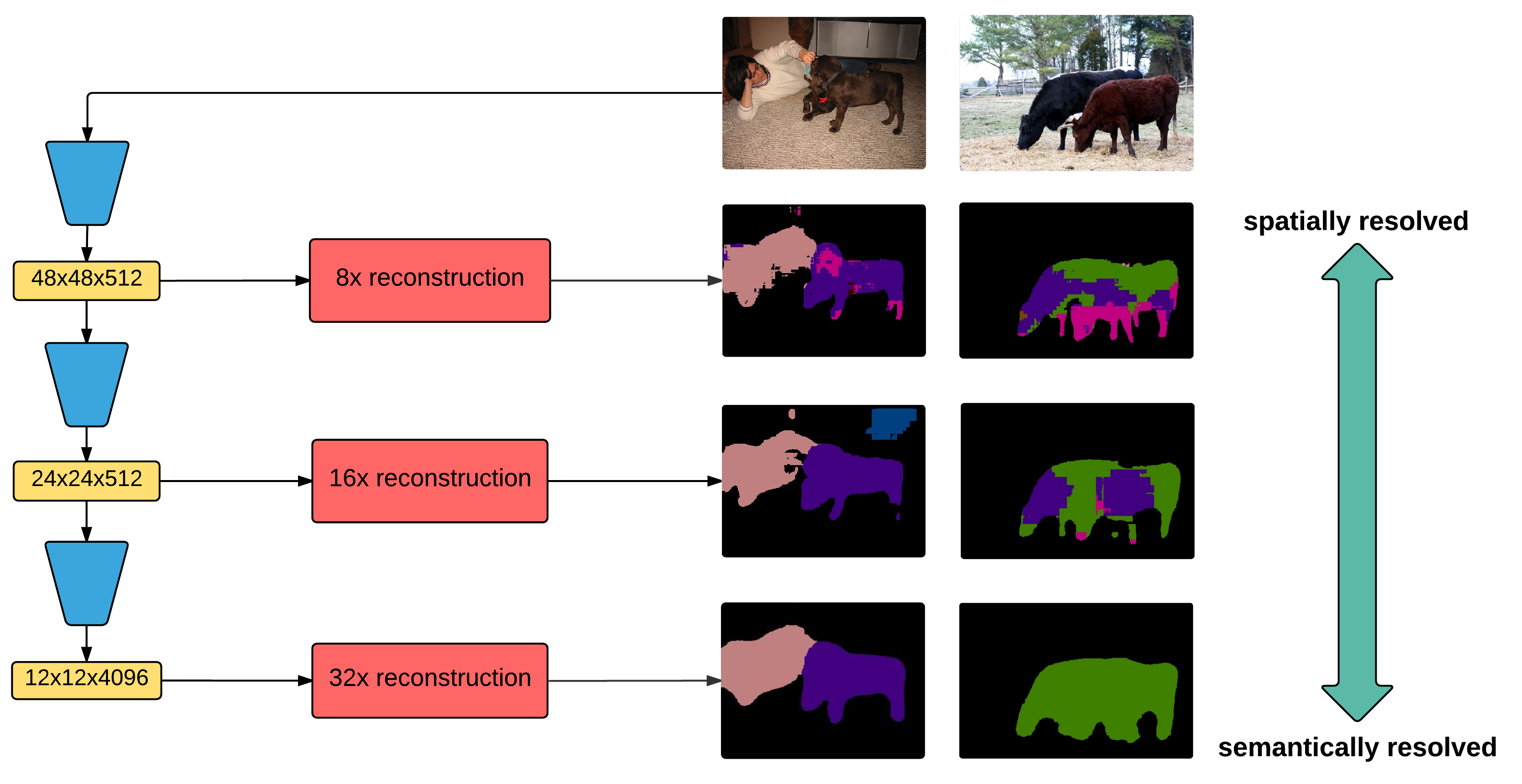}
\end{center}
\caption{In this paper, we explore the trade-off between spatial and semantic
accuracy within CNN feature hierarchies. Such hierarchies generally follow a
spatial-semantic uncertainty principle in which high levels of the hierarchy
make accurate semantic predictions but are poorly localized in space while at
low levels, boundaries are precise but labels are noisy.  We develop
reconstruction techniques for increasing spatial accuracy at a given level and
refinement techniques for fusing multiple levels that limit these tradeoffs and
produce improved semantic segmentations.
}
\label{fig:splash}
\vspace{-0.15in}
\end{figure}

Deep convolutional neural networks (CNNs) have proven highly effective at
semantic segmentation due to the capacity of discriminatively pre-trained
feature hierarchies to robustly represent and recognize objects and materials.
As a result, CNNs have significantly outperformed previous approaches (e.g.,
\cite{textonboost,carreirasiminchisescupami2012,carreiraijcv2012}) that relied
on hand-designed features and recognizers trained from scratch.  A key
difficulty in the adaption of CNN features to segmentation is that feature
pooling layers, which introduce invariance to spatial deformations required
for robust recognition, result in high-level representations with reduced
spatial resolution.  In this paper, we investigate this {\em spatial-semantic
uncertainty principle} for CNN hierarchies (see Fig.\ref{fig:splash}) and
introduce two techniques that yield substantially improved segmentations.

\begin{figure}[ht]
\newcommand{\figname}{2010_001069}
\newcommand{\clname}{bird}
\newcommand{\dirname}{figs/upsampling/}
\newcommand{\imw}{0.12}
\begin{tabular}{cc}
	\begin{tabular}{cc}
		\includegraphics[ height=0.32\linewidth, keepaspectratio,]{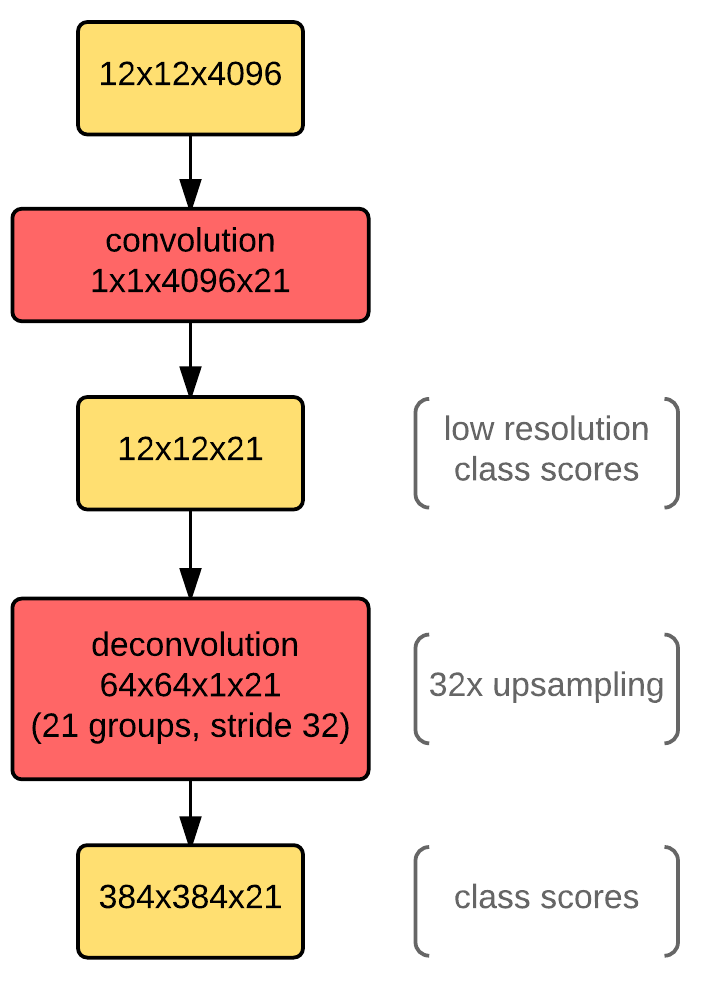}&
		\hspace{0.3cm}
		\includegraphics[ height=0.32\linewidth, keepaspectratio,]{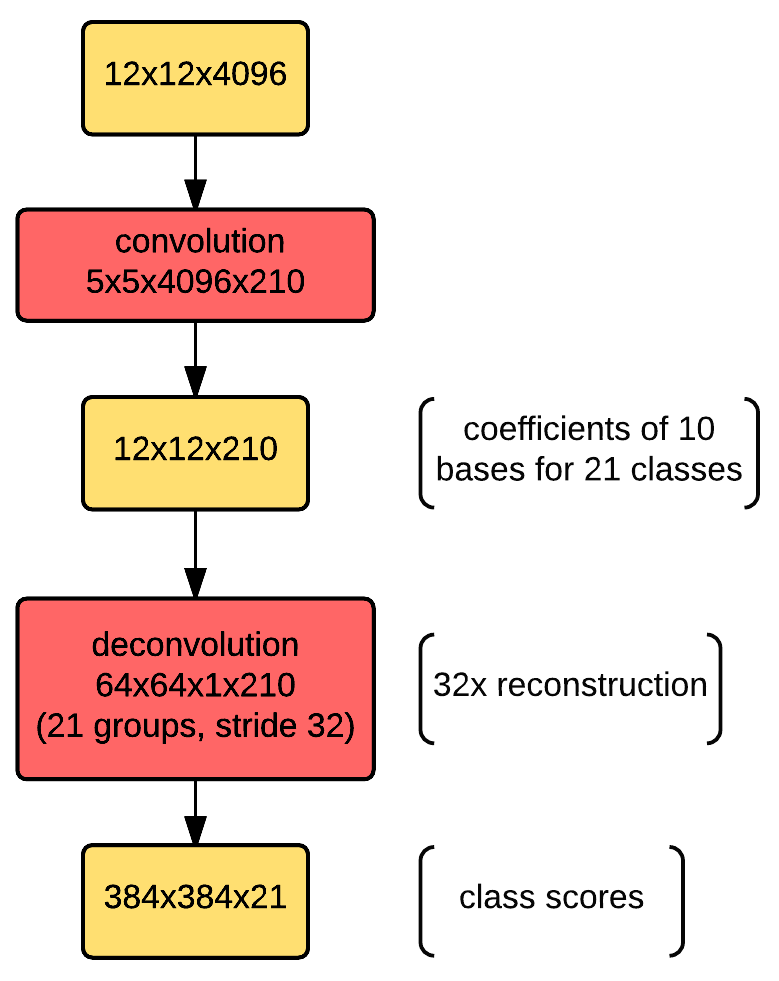}\\
	\end{tabular}&
	\hspace{0.5cm}
	\begin{tabular}{>{\tiny}c>{\tiny}c>{\tiny}c}
		\includegraphics[ width=\imw\linewidth, keepaspectratio,]{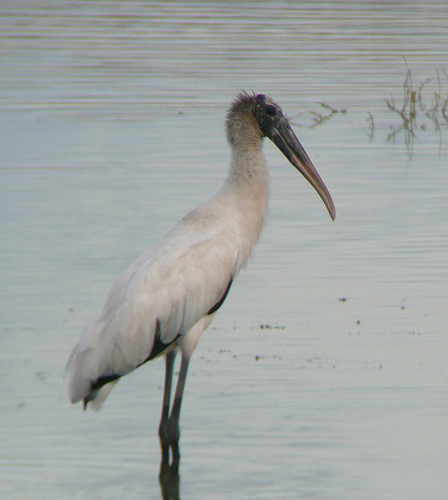}&
		\includegraphics[ width=\imw\linewidth, keepaspectratio,]{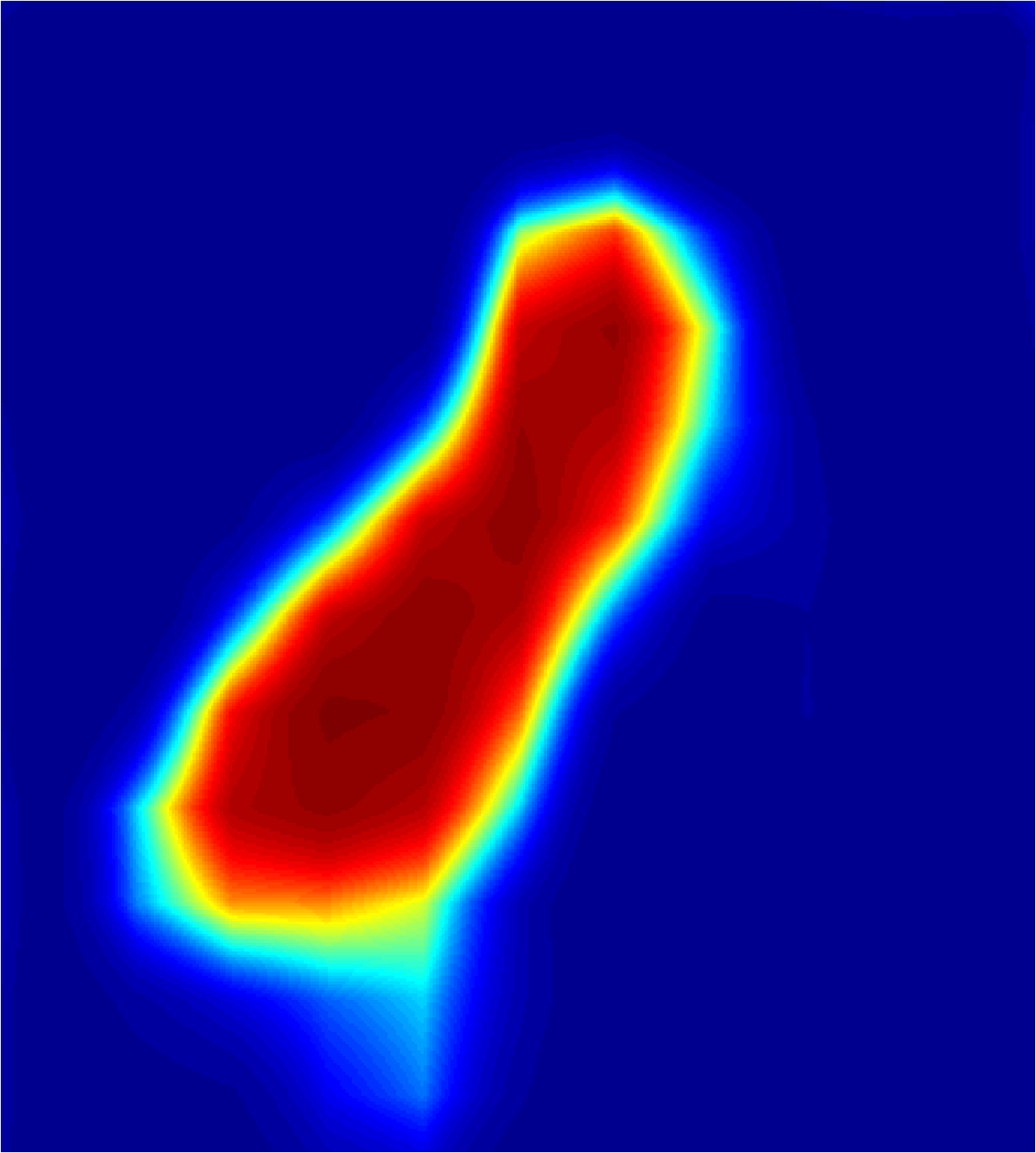}&
		\includegraphics[ width=\imw\linewidth, keepaspectratio,]{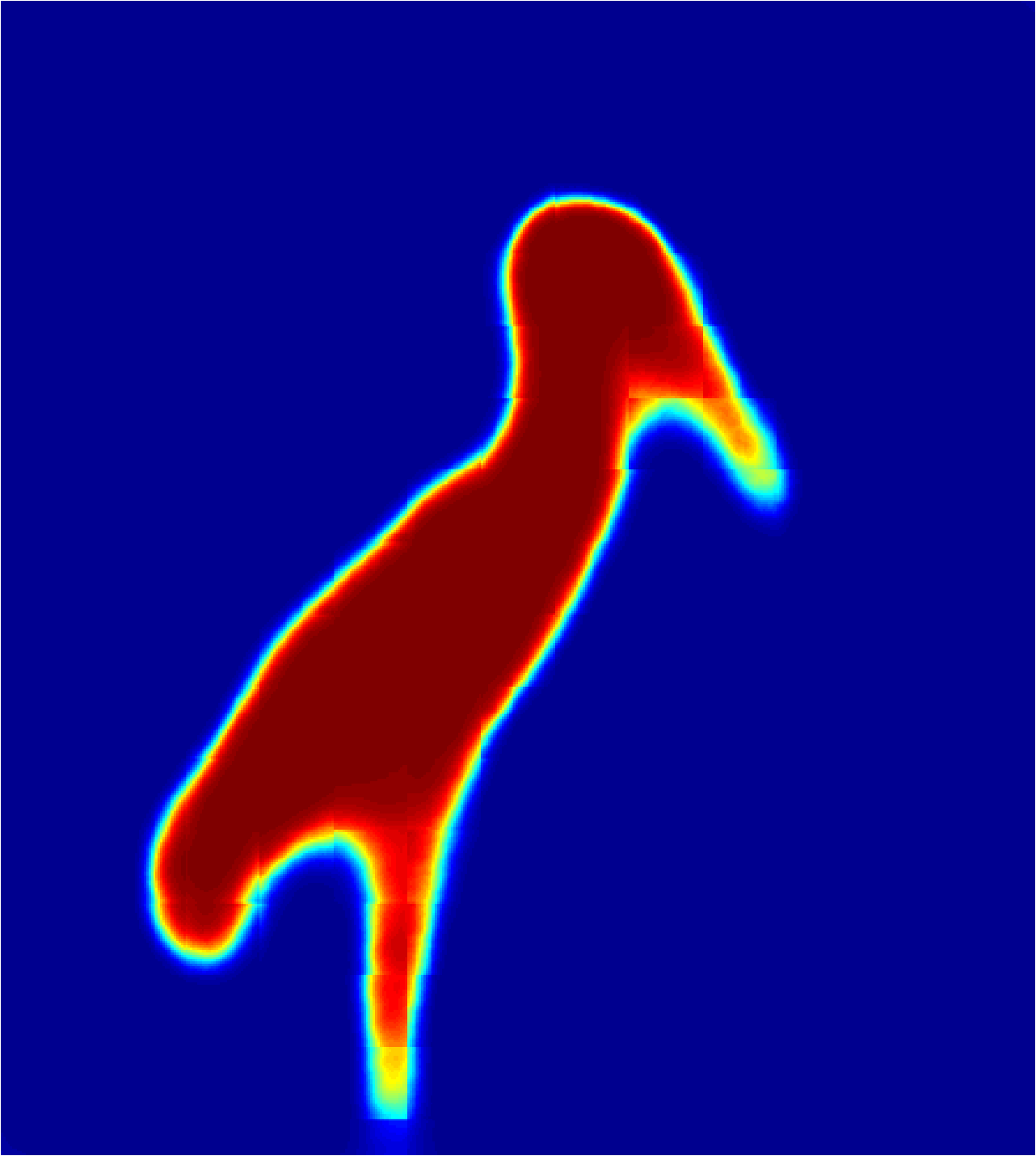} \\
		\includegraphics[ width=\imw\linewidth, keepaspectratio,]{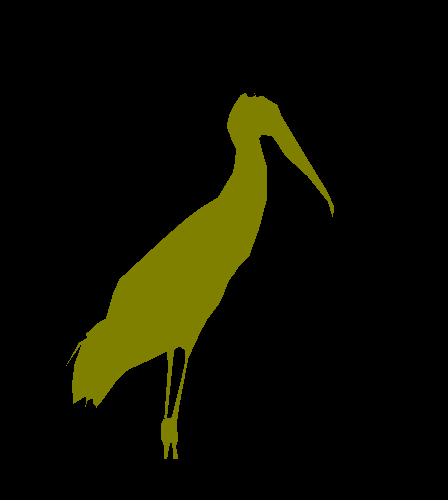} &
		\includegraphics[ width=\imw\linewidth, keepaspectratio,]{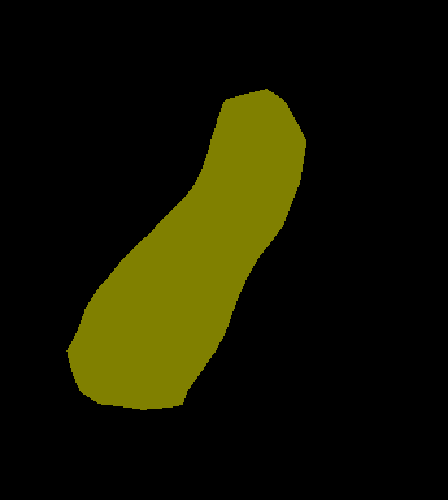} &
		\includegraphics[ width=\imw\linewidth, keepaspectratio,]{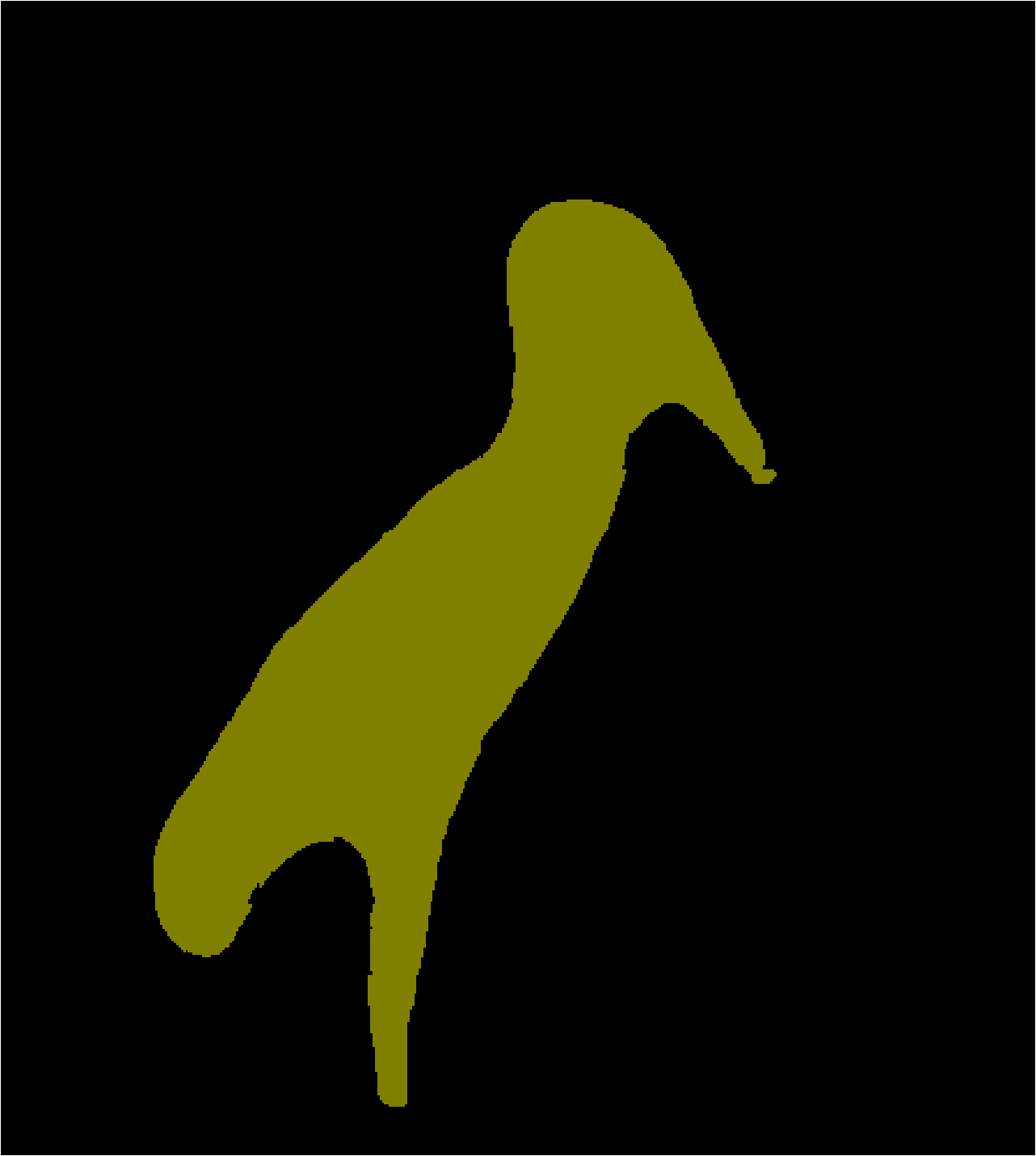} \\
    ground truth & upsampling & reconstruction\\
	\end{tabular}\\
        (a)&(b)\\
\end{tabular}
\caption{(a) Upsampling architecture for FCN32s network (left) and our 32x
reconstruction network (right).  (b) Example of Class conditional probability
maps and semantic segmentation predictions from FCN32s which performs
upsampling (middle) and our 32x reconstruction network (right).
}
\label{figure:upsampling}
\vspace{-0.15in}
\end{figure}

First, we tackle the question of how much spatial information is represented at
high levels of the feature hierarchy. A given spatial location in a
convolutional feature map corresponds to a large block of input pixels (and an
even larger ``receptive field'').  While max pooling in a single feature
channel clearly destroys spatial information in that channel, spatial filtering
prior to pooling introduces strong correlations across channels which could, in
principle, encode significant ``sub-pixel'' spatial information across the
high-dimensional vector of sparse activations.  We show that this is indeed the
case and demonstrate a simple approach to spatial decoding using a small set of
data-adapted basis functions that substantially improves over common
upsampling schemes (see Fig.~\ref{figure:upsampling}).

Second, having squeezed more spatial information from a given layer of the
hierarchy, we turn to the question of fusing predictions across layers. A
standard approach has been to either concatenate features (e.g.,
\cite{hariharan2015hypercolumns}) or linearly combine predictions (e.g.,
\cite{long2015fcn}). Concatenation is appealing but suffers from the high
dimensionality of the resulting features.  On the other hand, additive
combinations of predictions from multiple layers does not make good use of the
relative spatial-semantic content tradeoff.  High-resolution layers are shallow
with small receptive fields and hence yield inherently noisy predictions with
high pixel-wise loss.  As a result, we observe their contribution is
significantly down-weighted relative to low-resolution layers during linear
fusion and thus they have relatively little effect on final predictions.

\begin{figure}[htb]
\includegraphics[ width=\linewidth, keepaspectratio,]{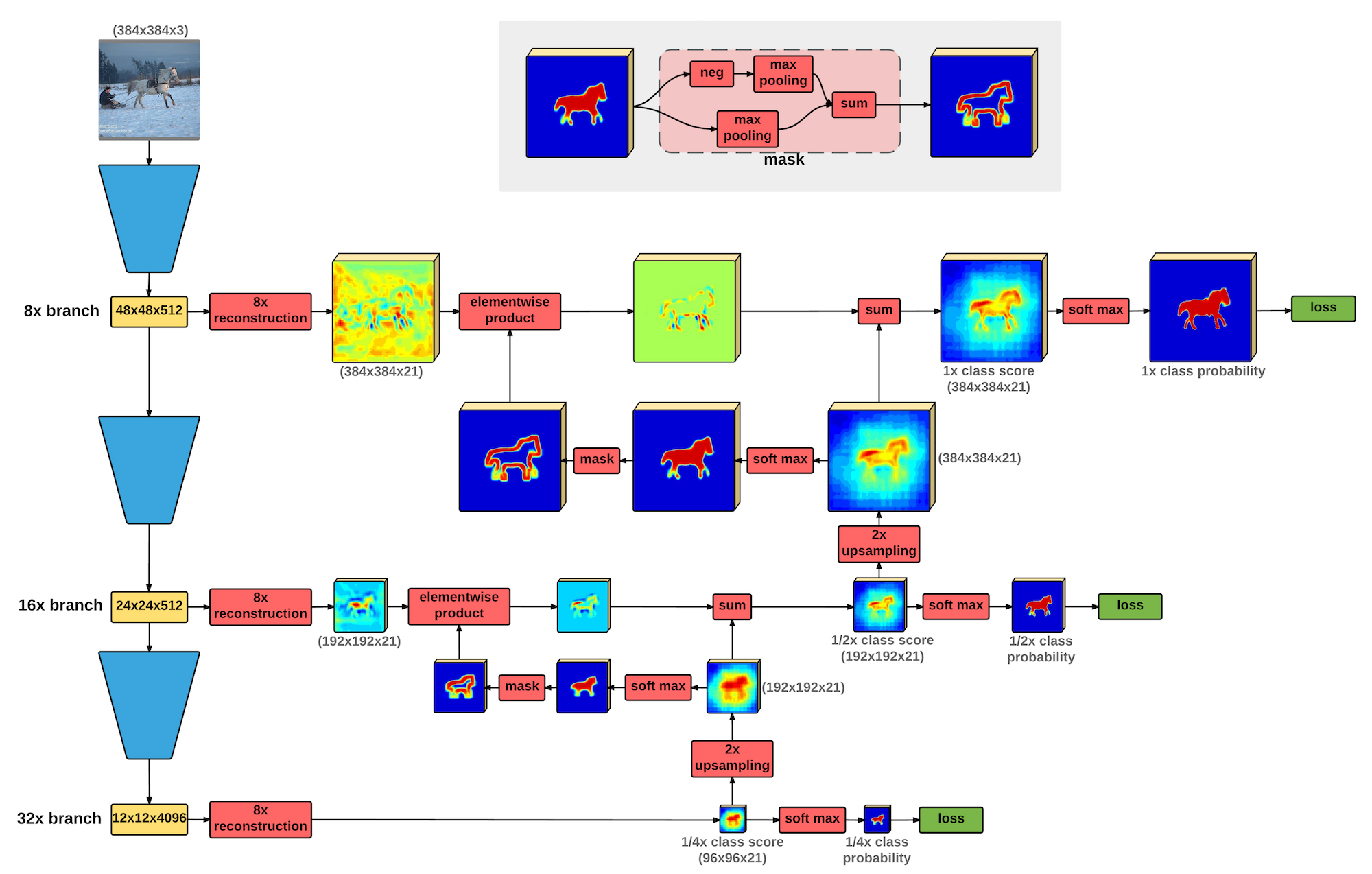}
\caption{Overview of our Laplacian pyramid reconstruction network architecture.
We use low-resolution feature maps in the CNN hierarchy to reconstruct a
coarse, low-frequency segmentation map and then refine this map by adding in
higher frequency details derived from higher-resolution feature maps. Boundary
masking (inset) suppresses the contribution of higher resolution layers in
areas where the segmentation is confident, allowing the reconstruction to focus
on predicting residual errors in uncertain areas (e.g., precisely localizing
object boundaries).  At each resolution layer, the reconstruction filters
perform the same amount of upsampling which depends on the number of layers
(e.g., our LRR-4x model utilizes 4x reconstruction on each of four branches).
Standard 2x bilinear upsampling is applied to each class score map before
combining it with higher resolution predictions.}
\label{figure:architecture}
\vspace{-0.15in}
\end{figure}

Inspired in part by recent work on residual networks
\cite{residualnetsarxiv2015,he2016identity}, we propose an architecture in
which predictions derived from high-resolution layers are only required to
correct residual errors in the low-resolution prediction.  Importantly, we use
multiplicative gating to avoid integrating (and hence penalizing) noisy
high-resolution outputs in regions where the low-resolution predictions are
confident about the semantic content.  We call our method {\em Laplacian
Pyramid Reconstruction and Refinement} (LRR) since the architecture uses a
Laplacian reconstruction pyramid~\cite{BurtAdelson} to fuse predictions.
Indeed, the class scores predicted at each level of our architecture typically
look like bandpass decomposition of the full resolution segmentation mask (see
Fig.~\ref{figure:architecture}).

\section{Related Work}
The inherent lack of spatial detail in CNN feature maps has been attacked using
a variety of techniques.  One insight is that spatial information lost during
max-pooling can in part be recovered by unpooling and
deconvolution~\cite{ZeilerICCV2011} providing a useful way to visualize input
dependency in feed-forward models~\cite{zeiler2014visualizing}.  This idea has
been developed using learned deconvolution filters to perform semantic
segmentation~\cite{noh2015deconvnet}.  However, the deeply stacked
deconvolutional output layers are difficult to train, requiring multi-stage
training and more complicated object proposal aggregation.

A second key insight is that while activation maps at lower-levels of the CNN
hierarchy lack object category specificity, they do contain higher spatial
resolution information.  Performing classification using a ``jet'' of feature
map responses aggregated across multiple layers has been successfully leveraged
for semantic segmentation~\cite{long2015fcn}, generic boundary detection
\cite{xie2015holistically}, simultaneous detection and segmentation
\cite{hariharan2015hypercolumns}, and scene recognition
\cite{SongfanYangRamanan2015}.  Our architecture shares the basic skip
connections of~\cite{long2015fcn} but uses multiplicative, confidence-weighted
gating when fusing predictions.

Our techniques are complementary to a range of other recent approaches that
incorporate object proposals \cite{dai2015boxsup,noh2015deconvnet}, attentional
scale selection mechanisms \cite{chen2015attention}, and conditional random
fields (CRF) \cite{chen2015deeplab,lin2016piecewise,liu2015parsing}.
CRF-based methods integrate CNN score-maps with pairwise features derived from
superpixels~\cite{Dai2014-eu,mostajabi2015zoomout} or generic boundary
detection~\cite{koki2016supbound,chen2015edgenet} to more precisely localize
segment boundaries.  We demonstrate that our architecture works well as a drop
in unary potential in fully connected CRFs \cite{koltun2011efficient} and would
likely further benefit from end-to-end training~\cite{zheng2015crfrnn}.

\section{Reconstruction with learned basis functions}
A standard approach to predicting pixel class labels is to use a linear
convolution to compute a low-resolution class score from the feature map and
then upsample the score map to the original image resolution. A bilinear kernel
is a suitable choice for this upsampling and has been used as a fixed filter
or an initialization for the upsampling filter
\cite{hariharan2015hypercolumns,long2015fcn,chen2015deeplab,zheng2015crfrnn,chen2015attention,dai2015boxsup,gidaris2015object}.
However, upsampling low-resolution class scores necessarily limits the amount of detail
in the resulting segmentation (see Fig.  \ref{figure:upsampling} (a)) and
discards any sub-pixel localization information that might be coded across the
many channels of the low-resolution feature map.  The simple fix of upsampling
the feature map prior to classification poses computational difficulties due to
the large number of feature channels (e.g. 4096). Furthermore, (bilinear)
upsampling commutes with 1x1 convolutions used for class prediction so
performing per-pixel linear classification on an upsampled feature map would
yield equivalent results unless additional rounds of (non-linear) filtering
were carried out on the high-resolution feature map.

To extract more detailed spatial information, we avoid immediately collapsing
the high-dimensional feature map down to low-resolution class scores.  Instead,
we express the spatial pattern of high-resolution scores using a linear
combination of high-resolution basis functions whose coefficients are predicted
from the feature map (see Fig.  \ref{figure:upsampling} (a)).  We term this
approach ``reconstruction'' to distinguish it from the standard
upsampling (although bilinear upsampling can clearly be seen as special case
with a single basis function).\\

\noindent{\bf Reconstruction by deconvolution:}
In our implementation, we tile the high-resolution score map with overlapping
basis functions (e.g., for 4x upsampled reconstruction we use basis functions
with an 8x8 pixel support and a stride of 4).  We use a convolutional layer to
predict $K$ basis coefficients for each of $C$ classes from the
high-dimensional, low-resolution feature map.  The group of coefficients for
each spatial location and class are then multiplied by the set of basis
function for the class and summed using a standard deconvolution (convolution
transpose) layer.

To write this explicitly, let $s$ denote the stride, $q_s(i) = \lfloor
\frac{i}{s} \rfloor$ denote the quotient, and $m_s(i) = i \mathop{mod} s$ the
remainder of $i$ by $s$.  The reconstruction layer that maps basis coefficients
$X \in \mathbb{R}^{H \times W \times K \times C}$ to class scores $Y \in
\mathbb{R}^{sH \times sW \times C}$ using basis functions $B \in \mathbb{R}^{2s
\times 2s \times K \times C}$ is given by:
\begin{align}
\nonumber Y_c[i,j] &= \sum_{k=0}^{K-1} \sum_{(u,v) \in \{0,1\}^2}  B_{k,c} 
    \left[m_s(i) + s\cdot u, m_s(j) + s \cdot v \right] 
    \cdot 
    X_{k,c}\left[q_s(i)-u,q_s(j)-v\right]
\end{align}
\noindent where $B_{k,c}$ contains the $k$-th basis function
for class $c$ with corresponding spatial weights $X_{k,c}$.  We assume
$X_{k,c}$ is zero padded and $Y_c$ is cropped appropriately.\\

\noindent{\bf Connection to spline interpolation:} 
We note that a classic approach to improving on bilinear interpolation is to
use a higher-order spline interpolant built from a standard set of
non-overlapping polynomial basis functions where the weights are determined
analytically to assure continuity between neighboring patches.  Our approach
using learned filters and basis functions makes minimal assumptions about
mapping from high dimensional activations to the coefficients $X$ but also
offers no guarantees on the continuity of $Y$.  We address this in part by
using larger filter kernels (i.e., $5\times5\times4096$) for predicting the
coefficients $X_{k,c}$ from the feature activations. This mimics the
computation used in spline interpolation of introducing linear dependencies
between neighboring basis weights and empirically improves continuity of the
output predictions.\\

\begin{figure}[t]
	\centering
    \setlength{\tabcolsep}{1pt}
    \renewcommand{\arraystretch}{0.5}
	\newcommand{\hb}{0.07}
	\newcommand{\colmap}{jet}
	\newcommand{\classnamea}{train}
	\begin{tabular}{>{\tiny}cccccccccccc}
	train&
	\includegraphics[ height=\hb\linewidth, keepaspectratio,]{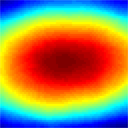}&
	\includegraphics[ height=\hb\linewidth, keepaspectratio,]{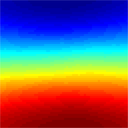}&
	\includegraphics[ height=\hb\linewidth, keepaspectratio,]{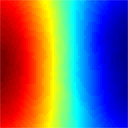}&
	\includegraphics[ height=\hb\linewidth, keepaspectratio,]{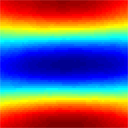}&
	\includegraphics[ height=\hb\linewidth, keepaspectratio,]{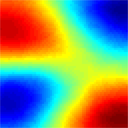}&
	\includegraphics[ height=\hb\linewidth, keepaspectratio,]{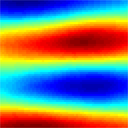}&
	\includegraphics[ height=\hb\linewidth, keepaspectratio,]{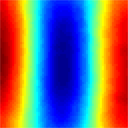}&
	\includegraphics[ height=\hb\linewidth, keepaspectratio,]{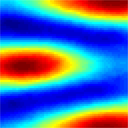}&
	\includegraphics[ height=\hb\linewidth, keepaspectratio,]{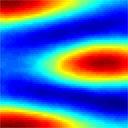}&
	\includegraphics[ height=\hb\linewidth, keepaspectratio,]{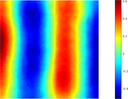}\\
	bird&
	\includegraphics[ height=\hb\linewidth, keepaspectratio,]{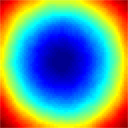}&
	\includegraphics[ height=\hb\linewidth, keepaspectratio,]{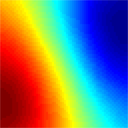}&
	\includegraphics[ height=\hb\linewidth, keepaspectratio,]{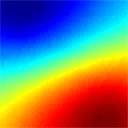}&
	\includegraphics[ height=\hb\linewidth, keepaspectratio,]{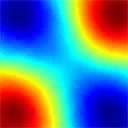}&
	\includegraphics[ height=\hb\linewidth, keepaspectratio,]{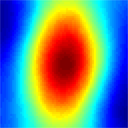}&
	\includegraphics[ height=\hb\linewidth, keepaspectratio,]{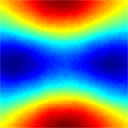}&
	\includegraphics[ height=\hb\linewidth, keepaspectratio,]{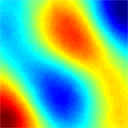}&
	\includegraphics[ height=\hb\linewidth, keepaspectratio,]{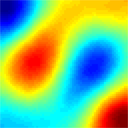}&
	\includegraphics[ height=\hb\linewidth, keepaspectratio,]{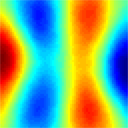}&
	\includegraphics[ height=\hb\linewidth, keepaspectratio,]{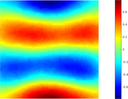}\\
	aeroplane&
	\includegraphics[ height=\hb\linewidth, keepaspectratio,]{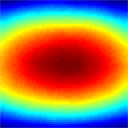}&
	\includegraphics[ height=\hb\linewidth, keepaspectratio,]{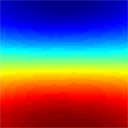}&
	\includegraphics[ height=\hb\linewidth, keepaspectratio,]{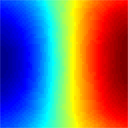}&
	\includegraphics[ height=\hb\linewidth, keepaspectratio,]{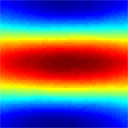}&
	\includegraphics[ height=\hb\linewidth, keepaspectratio,]{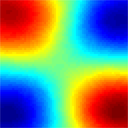}&
	\includegraphics[ height=\hb\linewidth, keepaspectratio,]{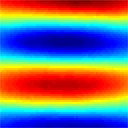}&
	\includegraphics[ height=\hb\linewidth, keepaspectratio,]{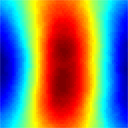}&
	\includegraphics[ height=\hb\linewidth, keepaspectratio,]{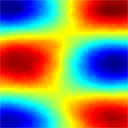}&
	\includegraphics[ height=\hb\linewidth, keepaspectratio,]{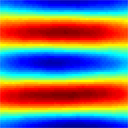}&
	\includegraphics[ height=\hb\linewidth, keepaspectratio,]{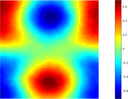}\\
	bottle&
	\includegraphics[ height=\hb\linewidth, keepaspectratio,]{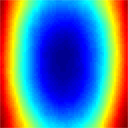}&
	\includegraphics[ height=\hb\linewidth, keepaspectratio,]{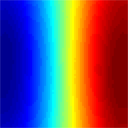}&
	\includegraphics[ height=\hb\linewidth, keepaspectratio,]{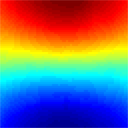}&
	\includegraphics[ height=\hb\linewidth, keepaspectratio,]{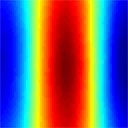}&
	\includegraphics[ height=\hb\linewidth, keepaspectratio,]{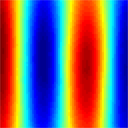}&
	\includegraphics[ height=\hb\linewidth, keepaspectratio,]{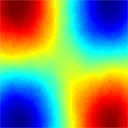}&
	\includegraphics[ height=\hb\linewidth, keepaspectratio,]{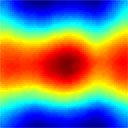}&
	\includegraphics[ height=\hb\linewidth, keepaspectratio,]{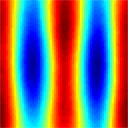}&
	\includegraphics[ height=\hb\linewidth, keepaspectratio,]{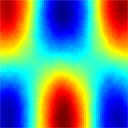}&
	\includegraphics[ height=\hb\linewidth, keepaspectratio,]{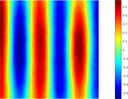}\\
	diningtable&
	\includegraphics[ height=\hb\linewidth, keepaspectratio,]{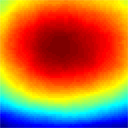}&
	\includegraphics[ height=\hb\linewidth, keepaspectratio,]{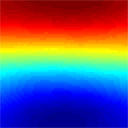}&
	\includegraphics[ height=\hb\linewidth, keepaspectratio,]{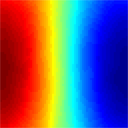}&
	\includegraphics[ height=\hb\linewidth, keepaspectratio,]{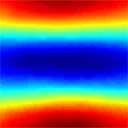}&
	\includegraphics[ height=\hb\linewidth, keepaspectratio,]{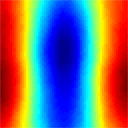}&
	\includegraphics[ height=\hb\linewidth, keepaspectratio,]{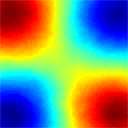}&
	\includegraphics[ height=\hb\linewidth, keepaspectratio,]{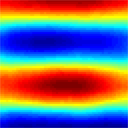}&
	\includegraphics[ height=\hb\linewidth, keepaspectratio,]{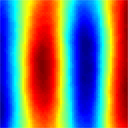}&
	\includegraphics[ height=\hb\linewidth, keepaspectratio,]{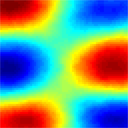}&
	\includegraphics[ height=\hb\linewidth, keepaspectratio,]{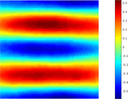}\\
	tvmonitor&
	\includegraphics[ height=\hb\linewidth, keepaspectratio,]{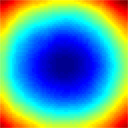}&
	\includegraphics[ height=\hb\linewidth, keepaspectratio,]{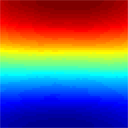}&
	\includegraphics[ height=\hb\linewidth, keepaspectratio,]{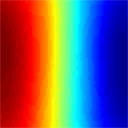}&
	\includegraphics[ height=\hb\linewidth, keepaspectratio,]{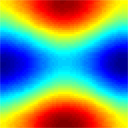}&
	\includegraphics[ height=\hb\linewidth, keepaspectratio,]{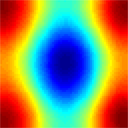}&
	\includegraphics[ height=\hb\linewidth, keepaspectratio,]{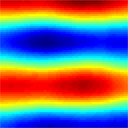}&
	\includegraphics[ height=\hb\linewidth, keepaspectratio,]{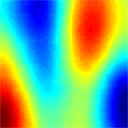}&
	\includegraphics[ height=\hb\linewidth, keepaspectratio,]{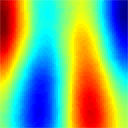}&
	\includegraphics[ height=\hb\linewidth, keepaspectratio,]{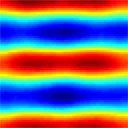}&
	\includegraphics[ height=\hb\linewidth, keepaspectratio,]{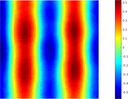}\\
	horse&
	\includegraphics[ height=\hb\linewidth, keepaspectratio,]{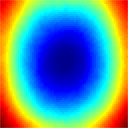}&
	\includegraphics[ height=\hb\linewidth, keepaspectratio,]{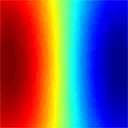}&
	\includegraphics[ height=\hb\linewidth, keepaspectratio,]{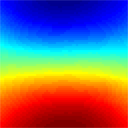}&
	\includegraphics[ height=\hb\linewidth, keepaspectratio,]{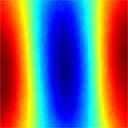}&
	\includegraphics[ height=\hb\linewidth, keepaspectratio,]{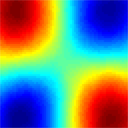}&
	\includegraphics[ height=\hb\linewidth, keepaspectratio,]{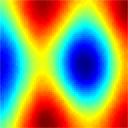}&
	\includegraphics[ height=\hb\linewidth, keepaspectratio,]{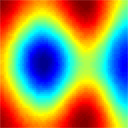}&
	\includegraphics[ height=\hb\linewidth, keepaspectratio,]{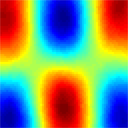}&
	\includegraphics[ height=\hb\linewidth, keepaspectratio,]{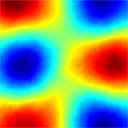}&
	\includegraphics[ height=\hb\linewidth, keepaspectratio,]{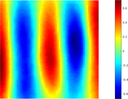}\\
	\end{tabular}
\caption{Category-specific basis functions for reconstruction are adapted to
modeling the shape of a given object class.  For example, airplane segments
tend to be elongated in the horizontal direction while bottles are elongated in
the vertical direction.}
\label{figure:bases}
\vspace{-0.15in}
\end{figure}

\noindent{\bf Learning basis functions:}
To leverage limited amounts of training data and speed up training, we
initialize the deconvolution layers with a meaningful set of filters estimated
by performing PCA on example segment patches.  For this purpose, we extract
$10000$ patches for each class from training data where each patch is of
size $32\times32$ and at least $2\%$ of the patch pixels are members of the
class.  We apply PCA on the extracted patches to compute a class specific set
of basis functions.  Example bases for different categories of PASCAL VOC
dataset are shown in Fig.  \ref{figure:bases}.  Interestingly, there is some
significant variation among classes due to different segment shape statistics.
We found it sufficient to initialize the reconstruction filters for different
levels of the reconstruction pyramid with the same basis set (downsampled as
needed). In both our model and the FCN bilinear upsampling model, we observed
that end-to-end training resulted in insignificant ($<10^{-7}$) changes to the
basis functions.

We experimented with varying the resolution and number of basis functions 
of our reconstruction layer built on top of the ImageNet-pretrained VGG-16
network. We found that $10$ functions sampled at a resolution of $8\times8$
were sufficient for accurate reconstruction of class score maps.  Models
trained with more than $10$ basis functions commonly predicted zero weight
coefficients for the higher-frequency basis functions.  This suggests some
limit to how much spatial information can be extracted from the low-res feature
map (i.e., roughly 3x more than bilinear). However, this estimate is only a
lower-bound since there are obvious limitations to how well we can fit the
model. Other generative architectures (e.g., using larger sparse dictionaries)
or additional information (e.g., max pooling ``switches'' in
deconvolution~\cite{ZeilerICCV2011}) may do even better.


\begin{figure}[ht!]
  \setlength{\tabcolsep}{1.4pt}
	\centering
    \setlength{\tabcolsep}{1pt}
    \renewcommand{\arraystretch}{0.5}
	\newcommand{\wa}{0.16}
	\newcommand{\figswodir}{figs/maskvsnot/wboundmaks/}
	\newcommand{\figswdir}{figs/maskvsnot/woboundmasks/}
	\begin{tabular}{>{\tiny}c>{\tiny}c>{\tiny}c>{\tiny}c>{\tiny}c>{\tiny}c}
	\includegraphics[ width=\wa\linewidth, keepaspectratio,]{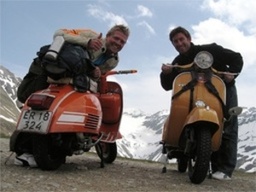}&
	\includegraphics[ width=\wa\linewidth, keepaspectratio,]{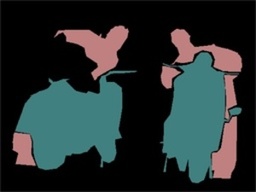}&
	\includegraphics[ width=\wa\linewidth, keepaspectratio,]{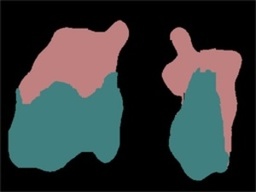}&
	\includegraphics[ width=\wa\linewidth, keepaspectratio,]{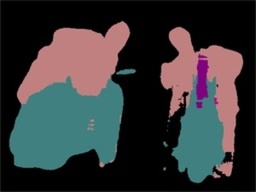}&
	\includegraphics[ width=\wa\linewidth, keepaspectratio,]{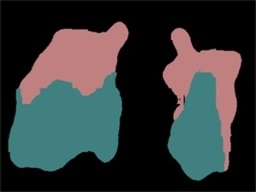}&
	\includegraphics[ width=\wa\linewidth, keepaspectratio,]{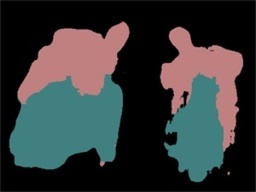}\\
	\includegraphics[ width=\wa\linewidth, keepaspectratio,]{\figswdir 54-input.jpg}&
	\includegraphics[ width=\wa\linewidth, keepaspectratio,]{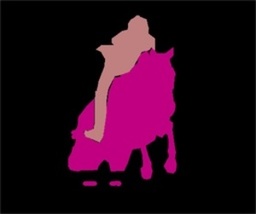}&
	\includegraphics[ width=\wa\linewidth, keepaspectratio,]{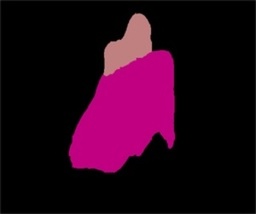}&
	\includegraphics[ width=\wa\linewidth, keepaspectratio,]{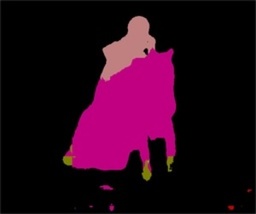}&
	\includegraphics[ width=\wa\linewidth, keepaspectratio,]{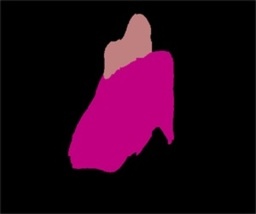}&
	\includegraphics[ width=\wa\linewidth, keepaspectratio,]{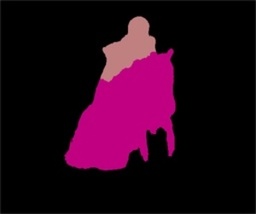}\\
	\includegraphics[ width=\wa\linewidth, keepaspectratio,]{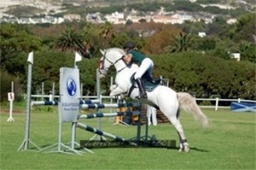}&
	\includegraphics[ width=\wa\linewidth, keepaspectratio,]{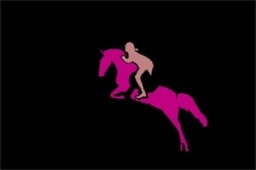}&
	\includegraphics[ width=\wa\linewidth, keepaspectratio,]{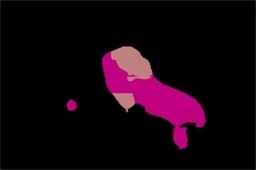}&
	\includegraphics[ width=\wa\linewidth, keepaspectratio,]{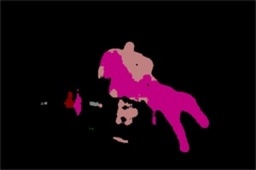}&
	\includegraphics[ width=\wa\linewidth, keepaspectratio,]{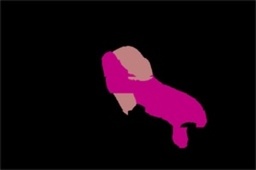}&
	\includegraphics[ width=\wa\linewidth, keepaspectratio,]{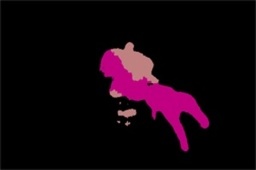}\\
	&ground-truth&32x unmask & 8x unmask & 32x masked &8x masked\\
	\end{tabular}
\caption{Visualization of segmentation results produced by our model with and
without boundary masking.  For each row, we show the input image, ground-truth
and the segmentation results of 32x and 8x layers of our model without masking
(middle) and with masking (right).  The segmentation results for 8x layer of
the model without masking has some noise not present in the 32x output.
Masking allows such noise to be repressed in regions where the 32x outputs have
high confidence.}
\label{figure:maskvsnot}
\vspace{-0.15in}
\end{figure}

\section{Laplacian Pyramid Refinement}

The basic intuition for our multi-resolution architecture comes from Burt and
Adelson's classic Laplacian Pyramid~\cite{BurtAdelson}, which decomposes an
image into disjoint frequency bands using an elegant recursive computation
(analysis) that produces appropriately down-sampled sub-bands such that the sum
of the resulting sub-bands (synthesis) perfectly reproduces the original image.
While the notion of frequency sub-bands is not appropriate for the non-linear
filtering performed by standard CNNs, casual inspection of the response of
individual activations to shifted input images reveals a power spectral density
whose high-frequency components decay with depth leaving primarily
low-frequency components (with a few high-frequency artifacts due to disjoint
bins used in pooling). This suggests the possibility that the standard CNN
architecture could be trained to serve the role of the analysis pyramid
(predicting sub-band coefficients) which could then be assembled using a
synthesis pyramid to estimate segmentations.

Figure \ref{figure:architecture} shows the overall architecture of our model.
Starting from the coarse scale ``low-frequency'' segmentation estimate, we
carry out a sequence of successive refinements, adding in information from
``higher-frequency'' sub-bands to improve the spatial fidelity of the resulting
segmentation masks. For example, since the 32x layer already captures the
coarse-scale support of the object, prediction from the 16x layer does not need
to include this information and can instead focus on adding finer scale
refinements of the segment boundary.
\footnote{Closely related architectures were used in \cite{denton2015deep} for
generative image synthesis where the output of a lower-resolution model was
used as input for a CNN which predicted an additive refinement, and in
\cite{piotrCVPR2016}, where fusing and refinement across levels was carried out
via concatenation followed by several convolution+ReLU layers.}\\

\noindent{\bf Boundary masking:}
In practice, simply upsampling and summing the outputs of the analysis layers
does not yield the desired effect. Unlike the Laplacian image analysis pyramid,
the high resolution feature maps of the CNN do not have the ``low-frequency''
content subtracted out.  As Fig.\ref{fig:splash} shows, high-resolution layers
still happily make ``low-frequency'' predictions (e.g., in the middle of a
large segment) even though they are often incorrect.  As a result, in an
architecture that simply sums together predictions across layers, we found the
learned parameters tend to down-weight the contribution of high-resolution
predictions to the sum in order to limit the potentially disastrous effect of
these noisy predictions.  However, this hampers the ability of the
high-resolution predictions to significantly refine the segmentation in areas
containing high-frequency content (i.e., segment boundaries).

To remedy this, we introduce a masking step that serves to explicitly subtract
out the ``low-frequency'' content from the high-resolution signal.  This takes
the form of a multiplicative gating that prevents the high-resolution
predictions from contributing to the final response in regions where
lower-resolution predictions are confident. The inset in
Fig.\ref{figure:architecture} shows how this boundary mask is computed by using
a max pooling operation to dilate the confident foreground and background
predictions and taking their difference to isolate the boundary.  The size of
this dilation (pooling size) is tied to the amount of upsampling between
successive layers of the pyramid, and hence fixed at 9 pixels in our implementation.

\begin{figure}[t]
\begin{center}
{
\scriptsize
\begin{tabular}{l|ccc}
{\bf VOC 2011-val} &pixel acc. & mean acc.     & mean IoU      \\ \hline
FCN-32s			          &89.1\%     & 73.3\%        & 59.4\%       \\
FCN-16s                           &90.0\%     & 75.7\%        & 62.4\%       \\
FCN-8s                            &90.3\%     & 75.9\%        & 62.7\%       \\ \hline
LRR-32x (w/o aug)                 &90.7\%    & 78.9\%       & 64.1\%      \\ \hline
LRR-32x                      &91.5\%    & 81.6\%       & 66.8\%      \\
LRR-16x                      &91.8\%    & 81.6\%       & 67.8\%      \\ 
LRR-8x                       &92.4\%    & 83.2\%       & 69.5\%      \\ 
LRR-4x                       &92.2\%    & 83.7\%       & 69.0\%      \\ \hline
LRR-4x-ms                    &92.8\%    & 84.6\%       & 71.4\%      \\ 
\end{tabular}
}
\end{center}
\caption{Comparison of our segment reconstruction model, LRR (without boundary
masking) and the baseline FCN model\cite{long2015fcn} which uses upsampling.  We
find consistent benefits from using a higher-dimensional reconstruction basis
rather than upsampling class prediction maps.  We also see improved performance
from using multi-scale training augmentation, fusing multiple feature maps, and
running on multiple scales at test time.  Note that the performance benefit of
fusing multiple resolution feature maps diminishes with no gain or even
decrease performance from adding in the 4x layer. Boundary masking (cf.
Fig.\ref{table:maskingresult}) allows for much better utilization of these fine
scale features.}
\label{table:fcnVsUs}
\vspace{-0.15in}
\end{figure}

\begin{figure}[t]
{
\scriptsize
\begin{center}
\begin{tabular}{l||c|c|c|c|c|c|c}
&\multicolumn{3}{c|}{VOC} &\multicolumn{4}{c}{VOC+COCO}  \\ \cline{2-8}
&\multicolumn{3}{c|}{VGG-16} &\multicolumn{3}{c|}{VGG-16}  &\multicolumn{1}{c}{ResNet} \\ \cline{2-8}
{\bf VOC 2011-val} & unmasked & masked &masked+DE  &umasked &masked & masked+DE & masked+DE\\
\hline
LRR-4x(32x)             & 67.1\%  &67.0\%  &68.8\% &71.3\% &71.2\%  & 72.9\%  & 76.7\%\\
LRR-4x(16x)             & 68.6\%  &69.2\%  &70.0\% &72.1\% &72.4\%  & 73.9\%  & 78.0\%\\
LRR-4x(8x)              & 69.3\%  &70.3\%  &70.9\% &72.9\% &73.4\%  & 74.9\%  & 78.3\%\\ \hline
LRR-4x                  & 69.3\%  &70.5\%  &71.1\% &72.9\% &73.6\%  & 75.1\%  & 78.4\%\\
LRR-4x-ms               & 71.9\%  &73.0\%  &73.6\% &74.0\% &75.0\%  & 76.6\%  & 79.2\%\\
LRR-4x-ms-crf           & 73.2\%  &74.1\%  &74.6\% &75.0\% &76.1\%  & 77.5\%  & 79.9\%\\
\end{tabular}
\end{center}
}
\caption{Mean intersection-over-union (IoU) accuracy for intermediate outputs
at different levels of our Laplacian reconstruction architecture trained with
and without boundary masking (value in parentheses denotes an intermediate
output of the full model).   Masking allows us to squeeze additional gains out
of high-resolution feature maps by focusing only on low-confidence areas near
segment boundaries.  Adding dilation and erosion losses (DE) to the 32x branch
improves the accuracy of 32x predictions and as a result the overall performance.
Running the model at multiple scales and performing post-processing using a CRF
yielded further performance improvements.}
\label{table:maskingresult}
\vspace{-0.15in}
\end{figure}

\section{Experiments}
We now describe a number of diagnostic experiments carried out using the PASCAL
VOC \cite{everingham2015pascal} semantic segmentation dataset.  In these
experiments, models were trained on training/validation set split specified by
\cite{hariharan2011semantic} which includes 11287 training images and 736 held
out validation images from the PASCAL 2011 val set.  We focus primarily on the
average Intersection-over-Union (IoU) metric which generally provides a more
sensitive performance measure than per-pixel or per-class accuracy.  We conduct
diagnostic experiments on the model architecture using this validation data and
test our final model via submission to the PASCAL VOC 2012 test data server,
which benchmarks on an additional set of 1456 images.  We also report test
benchmark performance on the recently released Cityscapes
\cite{Cordts2016Cityscapes} dataset.

\subsection{Parameter Optimization}
We augment the layers of the ImageNet-pretrained VGG-16 network
\cite{simonyan2014very} or ResNet-101 \cite{residualnetsarxiv2015} with our LRR
architecture and fine-tune all layers via back-propagation.  All models were
trained and tested with Matconvnet \cite{vedaldi15matconvnet} on a single
NVIDIA GPU. We use standard stochastic gradient descent with batch size of 20,
momentum of 0.9 and weight decay of 0.0005. The models and code are available
at {\tt https://github.com/golnazghiasi/LRR}.\\

\noindent{\bf Stage-wise training:}
Our 32x branch predicts a coarse semantic segmentation for the input image
while the other branches add in details to the segmentation prediction. Thus
16x, 8x and 4x branches are dependent on 32x branch prediction and their task
of adding details is meaningful only when 32x segmentation predictions are
good. As a result we first optimize the model with only 32x loss and then add
in connections to the other layers and continue to fine tune. At each layer we
use a pixel-wise softmax log loss defined at a lower image resolution and use
down-sampled ground-truth segmentations for training.  For example, in LRR-4x
the loss is defined at $1/8$, $1/4$, $1/2$ and full image resolution for the
32x, 16x, 8x and 4x branches, respectively.\\

\noindent{\bf Dilation erosion objectives:}
We found that augmenting the model with branches to predict dilated and eroded
class segments in addition of the original segments helps guide the model in
predicting more accurate segmentation.  For each training example and class, we
compute a binary segmentation using the ground-truth and then compute its
dilation and erosion using a disk with radius of $32$ pixels.  Since dilated
segments of different classes are not mutually exclusive, a k-way soft-max is
not appropriate so we use logistic loss instead.  We add these Dilation and
Erosion (DE) losses to the 32x branch (at $1/8$ resolution) when training
LRR-4x. Adding these losses increased mean IoU of the 32x branch predictions
from $71.2\%$ to  $72.9\%$ and also the overall multi-scale accuracy
from $75.0\%$ to $76.6$ (see Fig.\ref{table:maskingresult}, built on VGG-16 and
trained on VOC+COCO).\\

\begin{figure}[t]
	\centering
	\includegraphics[ width=0.46\linewidth, keepaspectratio,]{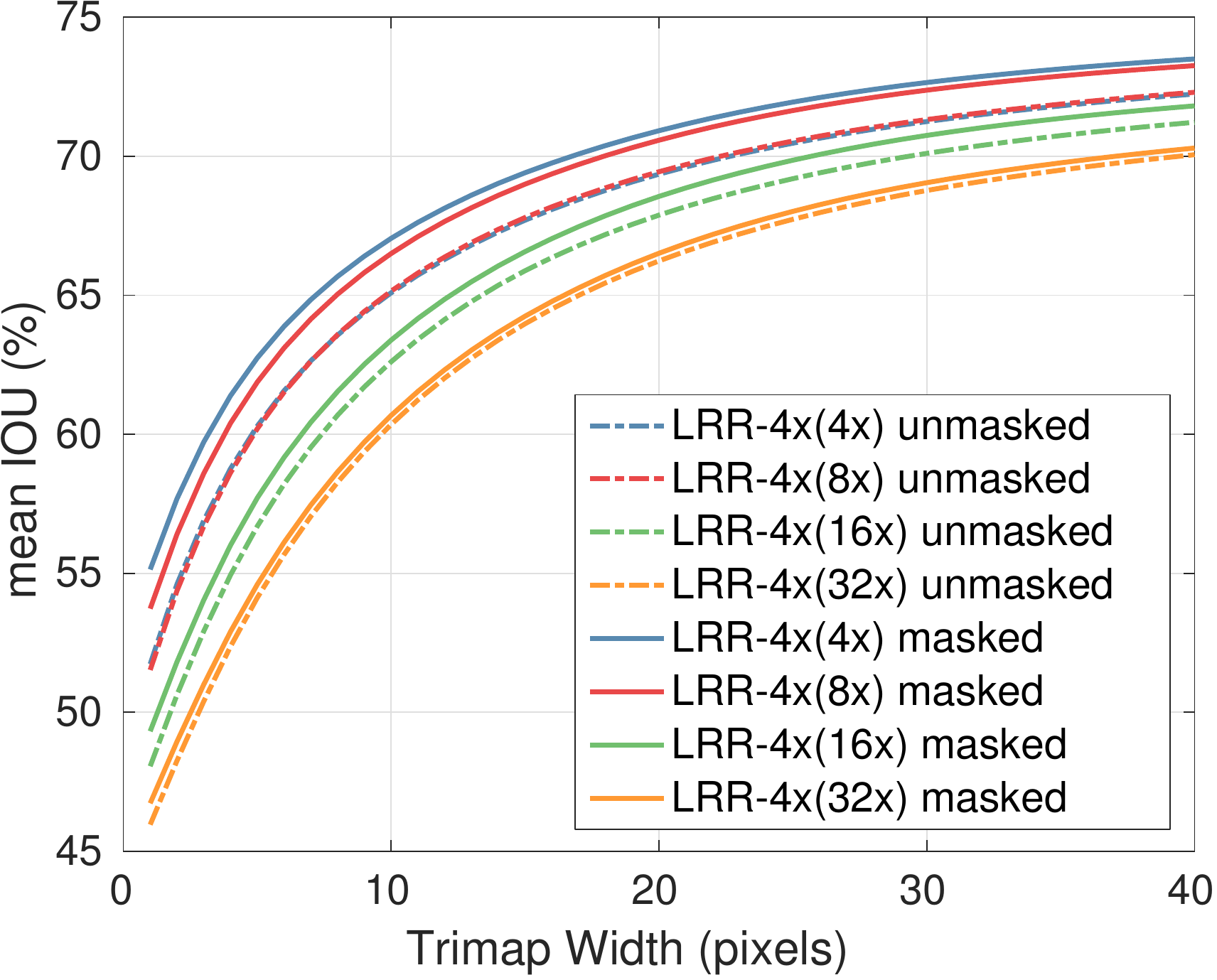}
	\includegraphics[ width=0.46\linewidth, keepaspectratio,]{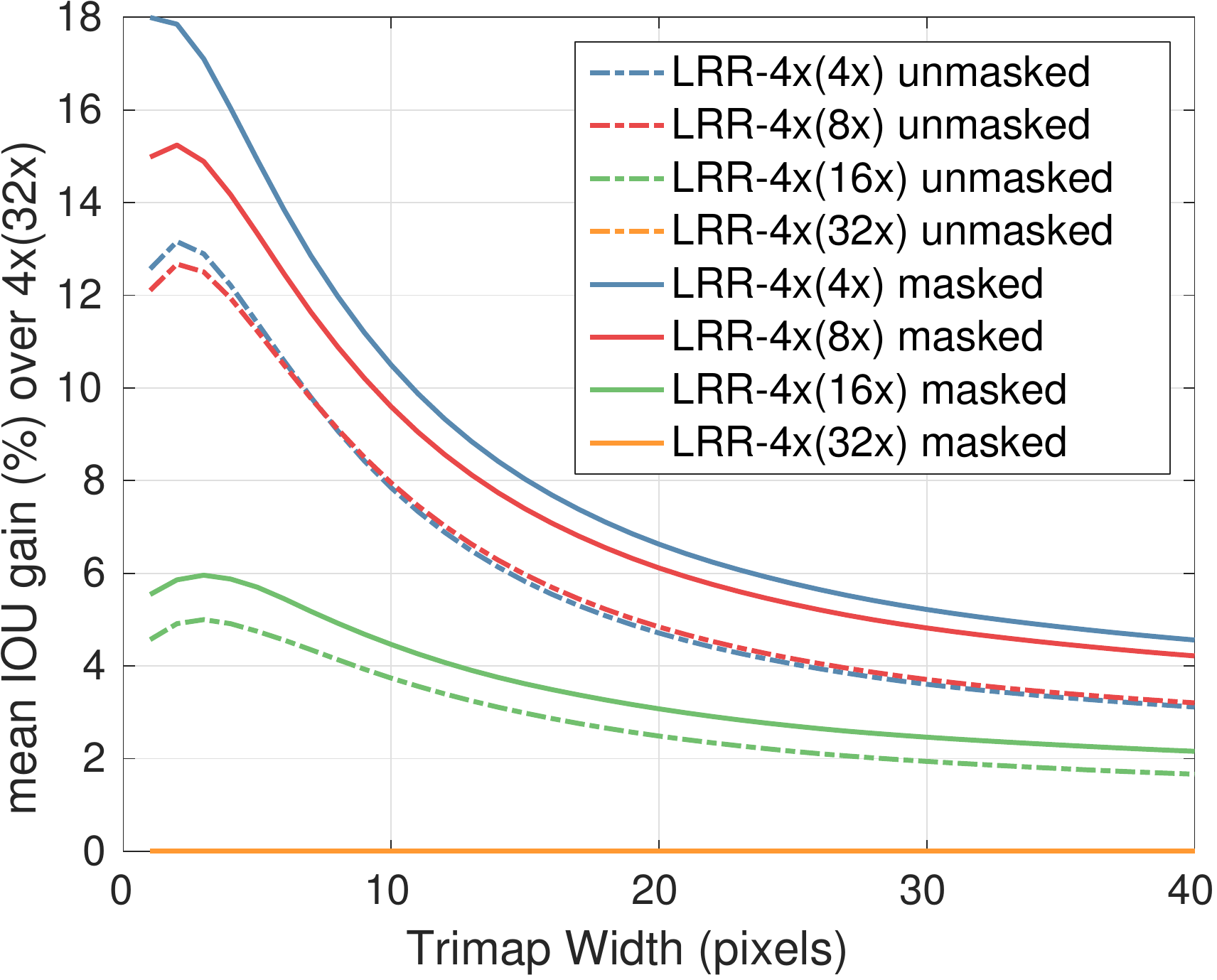}
\caption{ The benefit of Laplacian pyramid boundary refinement becomes even
more apparent when focusing on performance near object boundaries.  Plots show
segmentation performance within a thin band around the ground-truth object
boundaries for intermediate predictions at different levels of our
reconstruction pyramid. (right) Measuring accuracy or mean IoU relative to the
baseline 32x output shows that the high-resolution feature maps are most
valuable near object boundaries while masking improves performance both near
and far from boundaries.}
\label{figure:trimap}
\vspace{-0.15in}
\end{figure}

\noindent{\bf Multi-scale Data Augmentation:} \label{sec:scale-aug}
We augmented the training data with multiple scaled versions of each training
examples.  We randomly select an image size between 288 to 704 for each batch
and then scale training examples of that batch to the selected size. When the
selected size is larger than 384, we crop a window with size of 384$\times$384
from the scaled image.  This augmentation is helpful in improving the accuracy
of the model and increased mean IoU of our 32x model from 64.07\% to 66.81\% on
the validation data (see Fig.\ref{table:fcnVsUs}).

\subsection{Reconstruction vs Upsampling}
To isolate the effectiveness of our proposed reconstruction method relative to
simple upsampling, we compare the performance of our model without masking to
the fully convolutional net (FCN) of \cite{long2015fcn}.  For this experiment,
we trained our model without scale augmentation using exactly same training
data used for training the FCN models.  We observed significant improvement
over upsampling using reconstruction with 10 basis filters.  Our 32x
reconstruction model (w/o aug) achieved a mean IoU of 64.1\% while FCN-32s and
FCN-8s had a mean IoU of 59.4\% and 62.7\%, respectively (Fig.~\ref{table:fcnVsUs}).

\subsection{Multiplicative Masking and Boundary Refinement}
We evaluated whether masking the contribution of high-resolution feature maps
based on the confidence of the lower-resolution predictions resulted in better
performance.  We anticipated that this multiplicative masking would serve to
remove noisy class predictions from high-resolution feature maps in
high-confidence interior regions while allowing refinement of segment
boundaries.  Fig.  \ref{figure:maskvsnot} demonstrates the qualitative effect
of boundary masking. While the prediction from the 32x branch is similar for
both models (relatively noise free), masking improves the 8x prediction
noticeably by removing small, incorrectly labeled segments while preserving
boundary fidelity.  We compute mean IoU benchmarks for different intermediate
outputs of our LRR-4x model trained with and without masking (Table
\ref{table:maskingresult}). Boundary masking yields about 1\% overall improvement
relative to the model without masking across all branches.\\

\noindent{\bf Evaluation near Object Boundaries:}
Our proposed model uses the higher resolution feature maps to refine the
segmentation in the regions close to the boundaries, resulting in a more
detailed segmentation (see Fig. \ref{figure:qualitative}).  However, boundaries
constitute a relatively small fraction of the total image pixels, limiting the
impact of these improvements on the overall IoU performance benchmark (see,
e.g. Fig. \ref{table:maskingresult}).  To better characterize performance
differences between models, we also computed mean IoU restricted to a narrow
band of pixels around the ground-truth boundaries. This partitioning into
figure/boundary/background is sometimes referred to as a tri-map in the matting
literature and has been previously utilized in analyzing semantic segmentation
performance~\cite{chen2015deeplab,kohli2009hierpotentials}.

Fig.~\ref{figure:trimap} shows the mean IoU of our LRR-4x as a function of the
width of the tri-map boundary zone.  We plot both the absolute performance and
performance relative to the low-resolution 32x output.  As the curves confirm,
adding in higher resolution feature maps results in the most performance gain
near object boundaries. Masking improves performance both near and far from
boundaries.  Near boundaries masking allows for the higher-resolution layers to
refine the boundary shape while far from boundaries the mask prevents those
high-resolution layers from corrupting accurate low-resolution predictions.

\subsection{CRF Post-processing}
To show our architecture can easily be integrated with CRF-based models, we
evaluated the use of our LRR model predictions as a unary potential in a
fully-connected CRF \cite{koltun2011efficient,chen2015edgenet}.  We resize each
input image to three different scales (1,0.8,0.6), apply the LRR model and then
compute the pixel-wise maximum of predicted class conditional probability maps.
Post-processing with the CRF yields small additional gains in performance.
Fig.~\ref{table:maskingresult} reports the mean IoU for our LRR-4x model
prediction when running at multiple scales and with the integration of the CRF.
Fusing multiple scales yields a noticeable improvement (between $1.1\%$ to $2.5\%$)
while the CRF gives an additional gain (between 0.9\% to 1.4\%).

\begin{figure}[t]
\begin{center}
\def\rot{\rotatebox}
\setlength{\tabcolsep}{0.3pt}
{\tiny
\begin{tabular}{l||c||c|c|c|c|c|c|c|c|c|c|c|c|c|c|c|c|c|c|c|c}
          & \rot{90}{mean} & \rot{90}{areo} & \rot{90}{bike} &  \rot{90}{bird} &  \rot{90}{boat} & \rot{90}{bottle} & \rot{90}{bus}  & \rot{90}{car}  & \rot{90}{cat}  & \rot{90}{chair} & \rot{90}{cow} & \rot{90}{table} & \rot{90}{dog}  & \rot{90}{horse} & \rot{90}{mbike} & \rot{90}{person} & \rot{90}{plant} & \rot{90}{sheep} & \rot{90}{sofa} & \rot{90}{train} & \rot{90}{tv}  \\
\hline
\multicolumn{22}{c}{Only using VOC training data} \\
\hline
FCN-8s\cite{long2015fcn}            & 62.2 & 76.8 & 34.2 & 68.9 & 49.4 & 60.3 & 75.3 & 74.7 & 77.6 & 21.4 & 62.5 & 46.8 & 71.8 & 63.9 & 76.5 & 73.9 & 45.2 & 72.4 & 37.4 & 70.9 & 55.1 \\
Hypercol\cite{hariharan2015hypercolumns}& 62.6 & 68.7 & 33.5 & 69.8 & 51.3 & 70.2 & 81.1 & 71.9 & 74.9 & 23.9 & 60.6 & 46.9 & 72.1 & 68.3 & 74.5 & 72.9 & 52.6 & 64.4 & 45.4 & 64.9 & 57.4 \\
Zoom-out\cite{mostajabi2015zoomout} & 69.6 & 85.6 & 37.3 & \bf{83.2} & 62.5 & 66.0 & 85.1 & 80.7 & 84.9 & 27.2 & 73.2 & 57.5 & 78.1 & 79.2 & 81.1 & 77.1 & 53.6 & 74.0 & 49.2 & 71.7 & 63.3 \\
EdgeNet\cite{chen2015edgenet}       & 71.2 & 83.6 & 35.8 & 82.4 & 63.1 & 68.9 & 86.2 & 79.6 & 84.7 & 31.8 & 74.2 & 61.1 & 79.6 & 76.6 & 83.2 & 80.9 & 58.3 & 82.6 & 49.1 & 74.8 & 65.1 \\
Attention\cite{chen2015attention}   & 71.5 & 86.0 & 38.8 & 78.2 & 63.1 & 70.2 & 89.6 & 84.1 & 82.9 & 29.4 & 75.2 & 58.7 & 79.3 & 78.4 & 83.9 & 80.3 & 53.5 & 82.6 & 51.5 & 79.2 & 64.2 \\
DeepLab\cite{chen2015deeplab}       & 71.6 & 84.4 & 54.5 & 81.5 & 63.6 & 65.9 & 85.1 & 79.1 & 83.4 & 30.7 & 74.1 & 59.8 & 79.0 & 76.1 & 83.2 & 80.8 & 59.7 & 82.2 & 50.4 & 73.1 & 63.7 \\
CRFRNN\cite{zheng2015crfrnn}       & 72.0 & 87.5 & 39.0 & 79.7 & 64.2 & 68.3 & 87.6 & 80.8 & 84.4 & 30.4 & 78.2 & 60.4 & 80.5 & 77.8 & 83.1 & 80.6 & 59.5 & 82.8 & 47.8 & 78.3 & 67.1 \\
DeconvN\cite{noh2015deconvnet}    & 72.5 & 89.9 & 39.3 & 79.7 & 63.9 & 68.2 & 87.4 & 81.2 & 86.1 & 28.5 & 77.0 & 62.0 & 79.0 & 80.3 & 83.6 & 80.2 & 58.8 & 83.4 & 54.3 & 80.7 & 65.0 \\
DPN \cite{liu2015parsing}           & 74.1 & 87.7 & \bf{59.4} & 78.4 & 64.9 & 70.3 & 89.3 & 83.5 & 86.1 & 31.7 & 79.9 & 62.6 & 81.9 & 80.0 & 83.5 & 82.3 & 60.5 & 83.2 & 53.4 & 77.9 & 65.0 \\
Adelaide\cite{lin2016piecewise}   & 75.3 & 90.6 & 37.6 & 80.0 & \bf{67.8} & \bf{74.4} & 92.0 & 85.2 & 86.2 & \bf{39.1} & 81.2 & 58.9 & 83.8 & 83.9 & 84.3 & \bf{84.8} & \bf{62.1} & 83.2 & \bf{58.2} & 80.8 & \bf{72.3} \\\hline
LRR                               &74.7	& 89.2 & 40.3 &	81.2 & 63.9	& 73.1 & 91.7 & 86.2 & 87.2 & 35.4 & 80.1 & 62.4 & 82.6	& 84.4 & 84.8 & 81.7 & 59.5 & 83.6 & 54.3 & 83.7 & 69.3\\
LRR-CRF                             & \bf{75.9} & \bf{91.8} & 41.0 & 83.0 & 62.3 & 74.3 & \bf{93.0} & \bf{86.8} & \bf{88.7} & 36.6 & \bf{81.8} & \bf{63.4} & \bf{84.7} & \bf{85.9} & \bf{85.1}& 83.1 & 62.0 & \bf{84.6} & 55.6 & \bf{84.9} & 70.0 \\
\hline
\multicolumn{22}{c}{} \\
\multicolumn{22}{c}{Using VOC and COCO training data} \\
\hline
EdgeNet\cite{chen2015edgenet}       & 73.6 & 88.3 & 37.0 & \bf{89.8} & 63.6 & 70.3 & 87.3 & 82.0 & 87.6 & 31.1 & 79.0 & 61.9 & 81.6 & 80.4 & 84.5 & 83.3 & 58.4 & 86.1 & 55.9 & 78.2 & 65.4 \\
CRFRNN\cite{zheng2015crfrnn}       & 74.7 & 90.4 & 55.3 & 88.7 & 68.4 & 69.8 & 88.3 & 82.4 & 85.1 & 32.6 & 78.5 & 64.4 & 79.6 & 81.9 & 86.4 & 81.8 & 58.6 & 82.4 & 53.5 & 77.4 & 70.1 \\
BoxSup\cite{dai2015boxsup}          & 75.2 & 89.8 & 38.0 & 89.2 & \bf{68.9} & 68.0 & 89.6 & 83.0 & 87.7 & 34.4 & 83.6 & 67.1 & 81.5 & 83.7 & 85.2 & 83.5 & 58.6 & 84.9 & 55.8 & 81.2 & 70.7 \\
SBound\cite{koki2016supbound}     & 75.7 & 90.3 & 37.9 & 89.6 & 67.8 & 74.6 & 89.3 & 84.1 & 89.1 & 35.8 & 83.6 & 66.2 & 82.9 & 81.7 & 85.6 & 84.6 & 60.3 & 84.8 & 60.7 & 78.3 & 68.3 \\
Attention\cite{chen2015attention}   & 76.3 & 93.2 & 41.7 & 88.0 & 61.7 & 74.9 & 92.9 & 84.5 & 90.4 & 33.0 & 82.8 & 63.2 & 84.5 & 85.0 & 87.2 & 85.7 & 60.5 & 87.7 & 57.8 & \bf{84.3} & 68.2\\
DPN \cite{liu2015parsing}           & 77.5 & 89.0 & \bf{61.6} & 87.7 & 66.8 & 74.7 & 91.2 & 84.3 & 87.6 & 36.5 & 86.3 & 66.1 & 84.4 & 87.8 & 85.6 & 85.4 & 63.6 & 87.3 & 61.3 & 79.4 & 66.4 \\
Adelaide\cite{lin2016piecewise}     & 77.8 & \bf{94.1} & 40.4 & 83.6 & 67.3 & \bf{75.6} & 93.4 & 84.4 & 88.7 & 41.6 & \bf{86.4} & 63.3 & 85.5 & \bf{89.3} & 85.6 & 86.0 & \bf{67.4} & \bf{90.1} & 62.6 & 80.9 & \bf{72.5} \\\hline
LRR                                 &77.9  & 91.4 & 43.2 & 87.9 & 64.5 & 75.0 & 93.1 & 86.7 & 90.6 & 42.4 & 82.9 & 68.1 & 85.2 & 87.8 & 88.6 & 86.4 & 65.4 & 85.0 & 62.2 & 83.3 & 71.6\\
LRR-CRF                             & \bf{78.7} & 93.2 & 44.2 & 89.4	& 65.4 & 74.9 & \bf{93.9} & \bf{87.0} & \bf{92.0} & \bf{42.9} & 83.7 & \bf{68.9} & \bf{86.5} & 88.0 & \bf{89.0} & \bf{87.2} & 67.3	& 85.6 & \bf{64.0} & 84.1 & 71.5 \\
\hline
\multicolumn{22}{c}{} \\
\multicolumn{22}{c}{ResNet + Using VOC and COCO training data} \\
\hline
DeepLab\cite{chen2016deeplab}       & \bf{79.7} & \bf{92.6} & \bf{60.4} & 91.6	& 63.4 & \bf{76.3} & 95.0 & 88.4 & \bf{92.6} & 32.7 & \bf{88.5} & 67.6 & \bf{89.6} & \bf{92.1} & 87.0 & \bf{87.4} & 63.3 & \bf{88.3} & \bf{60.0} & \bf{86.8} & 74.5 \\\hline
LRR                                 & 78.7 & 90.8 & 44.4 & 94.0 & \bf{65.8} & 75.8 & 94.4 & 88.6 & 91.4 & \bf{39.1} & 84.7 & 70.0 & 87.5 & 88.7 & 88.3 & 85.8 & 64.1 & 85.6 & 56.6 & 85.1 & 76.8 \\
LRR-CRF                             & 79.3 & 92.4 & 45.1 & \bf{94.6} & 65.2 & 75.8 & \bf{95.1} & \bf{89.1} & 92.3 & 39.0 & 85.7 & \bf{70.4} & 88.6 & 89.4 & \bf{88.6} & 86.6 & \bf{65.8} & 86.2 & 57.4 & 85.7 & \bf{77.3} \\
\hline
\end{tabular}
}
\end{center}
\caption{Per-class mean intersection-over-union (IoU) performance on PASCAL VOC
2012 segmentation challenge test data. We evaluate models trained using only
VOC training data as well as those trained with additional training data from
COCO.  We also separate out a high-performing variant built on the ResNet-101
architecture.
}
\label{table:voc12test}
\label{table:voc12}
\vspace{-0.15in}
\end{figure}

\subsection{Benchmark Performance}

\noindent{\bf PASCAL VOC Benchmark:}
As the Table \ref{table:voc12} indicates, the current top performing
models on PASCAL all use additional training data from the MS COCO
dataset \cite{coco2014}.  To compare our approach with these architectures, we
also pre-trained versions of our model on MS COCO.  We utilized the 20
categories in COCO that are also present in PASCAL VOC, treated annotated
objects from other categories as background, and only used images where at
least 0.02\% of the image contained PASCAL classes. This resulted in 97765 out
of 123287 images of COCO training and validation set.

Training was performed in two stages.  In the first stage, we trained LRR-32x
on VOC images and COCO images together.  Since, COCO segmentation annotations
are often coarser in comparison to VOC segmentation annotations, we did not use
COCO images for training the LRR-4x. In the second stage, we used only PASCAL
VOC images to further fine-tune the LRR-32x and then added in connections to
the 16x, 8x and 4x layers and continue to fine-tune.  We used the multi-scale
data augmentation described in section \ref{sec:scale-aug} for both stages.
Training on this additional data improved the mean IoU of our model from 74.6\%
to 77.5\% on PASCAL VOC 2011 validation set (see
Table~\ref{table:maskingresult}).\\

\begin{figure}[!t]
  \centering
  \tiny
  \begin{tabular}{cc}
  \begin{tabular}{l||c|c}
    \addtolength{\tabcolsep}{.7pt}
    			     	    & unmasked+DE & masked+DE\\
    \hline
    LRR-4x(32x)             &64.7\%       &64.7\%\\
    LRR-4x(16x)             &66.7\%       &67.1\%\\
    LRR-4x(8x)              &68.5\%       &69.3\%\\
    LRR-4x                  &68.9\%       &70.0\%\\
  \end{tabular}\quad\quad&
  \begin{tabular}{l || c | c | c | c}
    \addtolength{\tabcolsep}{.7pt}
    			& {IoU class} & {iIoU class} & {IoU cat} & {iIoU cat}\\
    \hline
    FCN-8s \cite{long2015fcn} 						& 65.3\% & 41.7\% & 85.7\% & 70.1\%\\
    CRF-RNN \cite{zheng2015crfrnn} 					& 62.5\% & 34.4\% & 82.7\% & 66.0\%\\
    Dilation10 \cite{Yu2015-yy} 					& 67.1\% & 42.0\% & 86.5\% & 71.1\%\\
    DPN \cite{liu2015parsing}                       & 66.8\% & 39.1\% & 86.0\% & 69.1\%\\
    Pixel-level Encoding \cite{uhrig2016pixel}      & 64.3\% & 41.6\% & 85.9\% & 73.9\%\\
    DeepLab(ResNet) \cite{chen2016deeplab}      & 70.4\% & 42.6\% & 86.4\% & 67.7\%\\
    Adelaide\_Context \cite{lin2016piecewise}       & \bf{71.6\%} & \bf{51.7\%} & 87.3\% & 74.1\%\\
    \hline
	LRR-4x(VGG16)                                  & 69.7\% & 48.0\% & \bf{88.2\%} & \bf{74.7\%}\\
  \end{tabular}\\
  (a)&(b)\\
  \end{tabular}
  \caption{(a) Mean intersection-over-union (IoU class) accuracy on Cityscapes validation set
	for intermediate outputs at different levels of our Laplacian reconstruction architecture
	trained with and without boundary masking.
	(b) Comparison of our model with state-of-the-art methods on the Cityscapes benchmark test set.}
  \label{tab:cityscapes_test}
\vspace{-0.15in}
\end{figure}

\noindent{\bf Cityscapes Benchmark:} The Cityscapes dataset \cite{Cordts2016Cityscapes} 
contains high quality pixel-level annotations of images collected in street scenes
from 50 different cities. The training, validation, and test sets contain 2975, 500,
and 1525 images respectively (we did not use coarse annotations).  This dataset
contains labels for 19 semantic classes belonging to 7 categories of ground,
construction, object, nature, sky, human, and vehicle.

\begin{figure}[hb]
        \setlength{\tabcolsep}{1pt}
        \renewcommand{\arraystretch}{0.3}
	\centering
	\newcommand{\wa}{0.158}
	\newcommand{\figsdir}{figs/qualitive_res/4x-bmask-ep12/}
	\newcommand{\figfcndir}{figs/qualitive_res/fcn8s/}
	\newcommand{\fileid}{13}
	\begin{tabular}{>{\tiny}c>{\tiny}c>{\tiny}c>{\tiny}c>{\tiny}c>{\tiny}c}
	\includegraphics[ width=\wa\linewidth, keepaspectratio,]{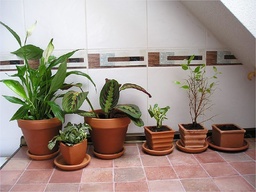}&%
	\renewcommand{\fileid}{13}%
	\includegraphics[ width=\wa\linewidth, keepaspectratio,]{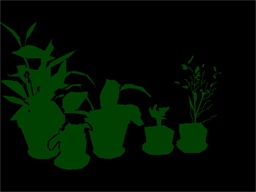}&%
	\renewcommand{\fileid}{13}%
	\includegraphics[ width=\wa\linewidth, keepaspectratio,]{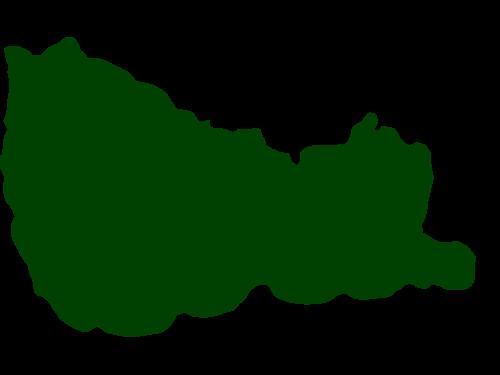}&%
	\renewcommand{\fileid}{13}%
	\includegraphics[ width=\wa\linewidth, keepaspectratio,]{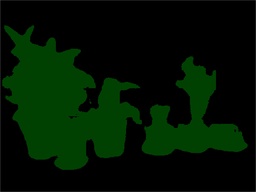}&%
	\renewcommand{\fileid}{13}%
	\includegraphics[ width=\wa\linewidth, keepaspectratio,]{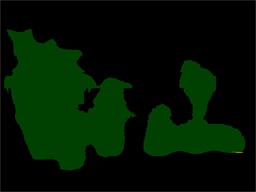}&%
	\renewcommand{\fileid}{13}%
	\includegraphics[ width=\wa\linewidth, keepaspectratio,]{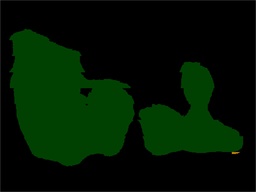}\\%
	\renewcommand{\fileid}{706}
	\includegraphics[ width=\wa\linewidth, keepaspectratio,]{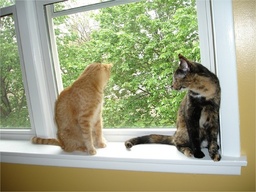}&
	\renewcommand{\fileid}{706}
	\includegraphics[ width=\wa\linewidth, keepaspectratio,]{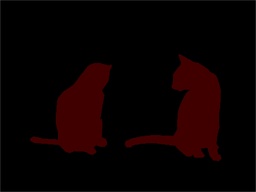}&%
	\renewcommand{\fileid}{706}
	\includegraphics[ width=\wa\linewidth, keepaspectratio,]{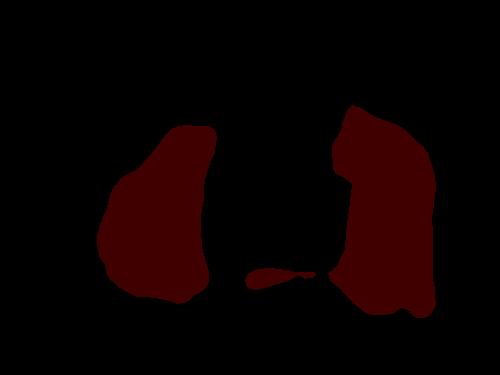}&%
	\renewcommand{\fileid}{706}
	\includegraphics[ width=\wa\linewidth, keepaspectratio,]{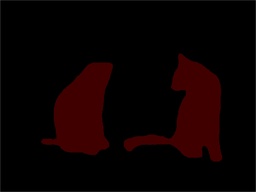}&%
	\renewcommand{\fileid}{706}
	\includegraphics[ width=\wa\linewidth, keepaspectratio,]{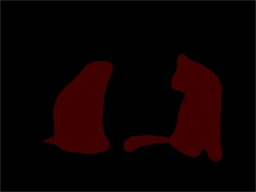}&%
	\renewcommand{\fileid}{706}
	\includegraphics[ width=\wa\linewidth, keepaspectratio,]{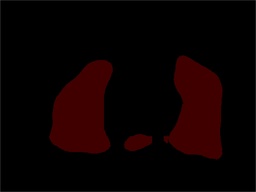}\\%
	\renewcommand{\fileid}{726}
	\includegraphics[ width=\wa\linewidth, keepaspectratio,]{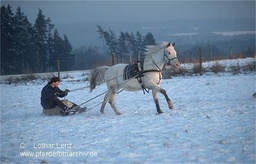}&%
	\renewcommand{\fileid}{726}
	\includegraphics[ width=\wa\linewidth, keepaspectratio,]{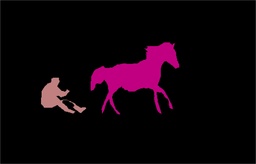}&%
	\renewcommand{\fileid}{726}
	\includegraphics[ width=\wa\linewidth, keepaspectratio,]{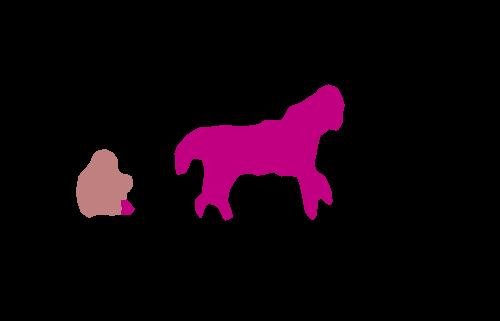}&%
	\renewcommand{\fileid}{726}
	\includegraphics[ width=\wa\linewidth, keepaspectratio,]{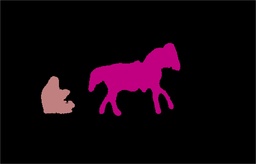}&%
	\renewcommand{\fileid}{726}
	\includegraphics[ width=\wa\linewidth, keepaspectratio,]{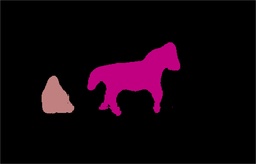}&%
	\renewcommand{\fileid}{726}
	\includegraphics[ width=\wa\linewidth, keepaspectratio,]{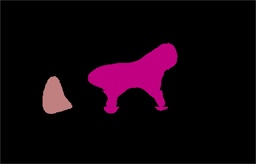}\\%
	\renewcommand{\fileid}{643}
	\includegraphics[ width=\wa\linewidth, keepaspectratio,]{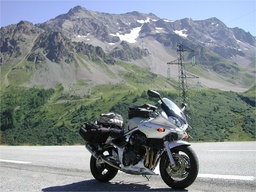}&%
	\renewcommand{\fileid}{643}
	\includegraphics[ width=\wa\linewidth, keepaspectratio,]{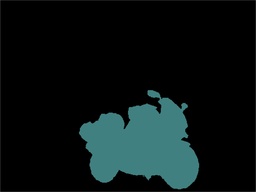}&%
	\renewcommand{\fileid}{643}
	\includegraphics[ width=\wa\linewidth, keepaspectratio,]{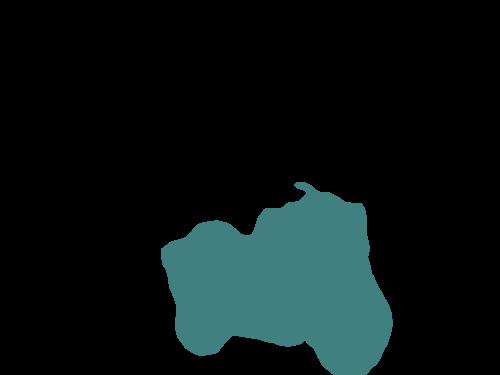}&%
	\renewcommand{\fileid}{643}
	\includegraphics[ width=\wa\linewidth, keepaspectratio,]{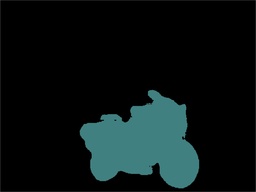}&%
	\renewcommand{\fileid}{643}
	\includegraphics[ width=\wa\linewidth, keepaspectratio,]{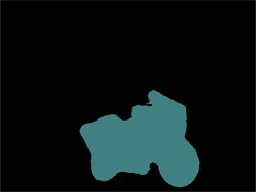}&%
	\renewcommand{\fileid}{643}
	\includegraphics[ width=\wa\linewidth, keepaspectratio,]{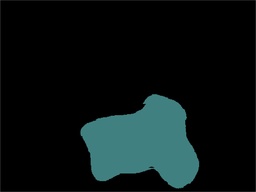}\\%
	\renewcommand{\fileid}{616}%
	\includegraphics[ width=\wa\linewidth, keepaspectratio,]{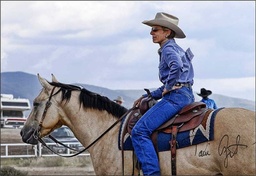}&%
	\renewcommand{\fileid}{616}%
	\includegraphics[ width=\wa\linewidth, keepaspectratio,]{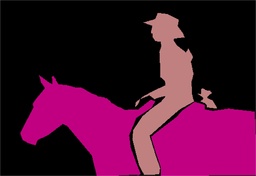}&%
	\renewcommand{\fileid}{616}%
	\includegraphics[ width=\wa\linewidth, keepaspectratio,]{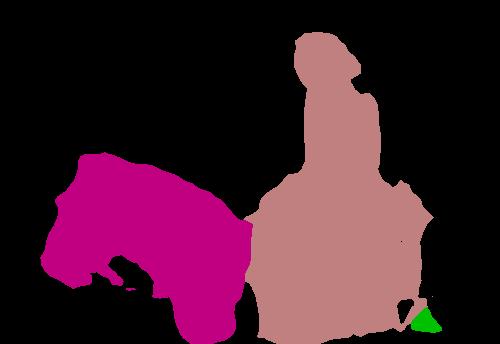}&%
	\renewcommand{\fileid}{616}%
	\includegraphics[ width=\wa\linewidth, keepaspectratio,]{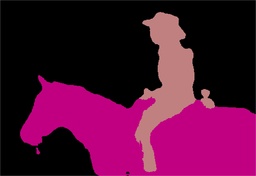}&%
	\renewcommand{\fileid}{616}%
	\includegraphics[ width=\wa\linewidth, keepaspectratio,]{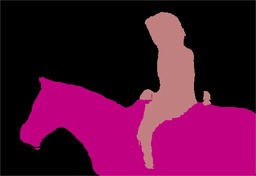}&%
	\renewcommand{\fileid}{616}%
	\includegraphics[ width=\wa\linewidth, keepaspectratio,]{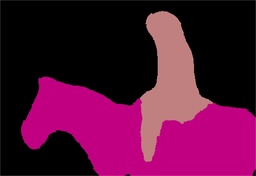}\\%
	\renewcommand{\fileid}{154}%
	\includegraphics[ width=\wa\linewidth, keepaspectratio,]{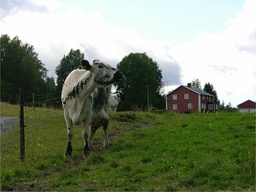}&%
	\renewcommand{\fileid}{154}%
	\includegraphics[ width=\wa\linewidth, keepaspectratio,]{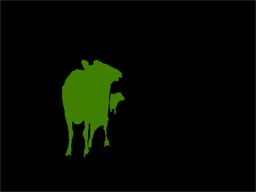}&%
	\renewcommand{\fileid}{154}%
	\includegraphics[ width=\wa\linewidth, keepaspectratio,]{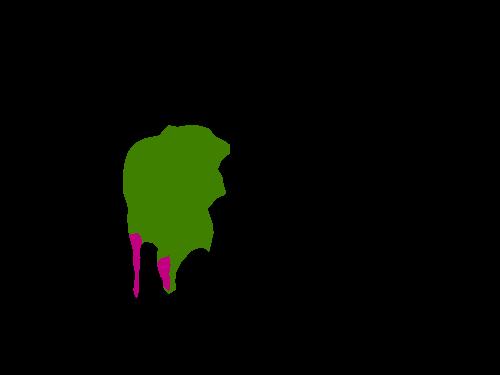}&%
	\renewcommand{\fileid}{154}%
	\includegraphics[ width=\wa\linewidth, keepaspectratio,]{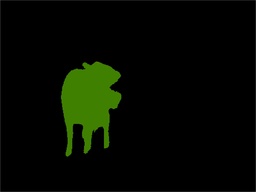}&%
	\renewcommand{\fileid}{154}%
	\includegraphics[ width=\wa\linewidth, keepaspectratio,]{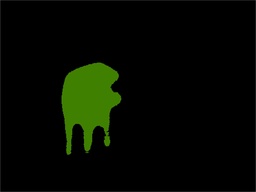}&%
	\renewcommand{\fileid}{154}%
	\includegraphics[ width=\wa\linewidth, keepaspectratio,]{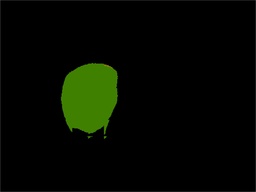}\\%
	&ground-truth&FCN-8s&LRR-4x (8x) & LRR-4x (16x) & LRR-4x (32x) \\
        \end{tabular}

        \vspace{0.2cm}
        \setlength{\tabcolsep}{1pt}
        \renewcommand{\arraystretch}{0.3}
	\centering
	\newcommand{\wb}{0.19}
	\newcommand{\figpdir}{figs/cityscape-qres/LRR/}
	\newcommand{\figgtdir}{figs/cityscape-qres/gt/}
	\newcommand{\fileaid}{frankfurt_000000_010763}
	\newcommand{\filebid}{frankfurt_000001_002646}
	\newcommand{\filecid}{frankfurt_000001_013710}
	\newcommand{\fileeid}{frankfurt_000001_015328}
	\newcommand{\filefid}{frankfurt_000001_030310}
	\newcommand{\filegid}{frankfurt_000001_048196}
	\newcommand{\filehid}{frankfurt_000001_052594}
	\newcommand{\fileiid}{frankfurt_000001_057954}
	\newcommand{\filejid}{lindau_000002_000019}
	\begin{tabular}{>{\tiny}c>{\tiny}c>{\tiny}c>{\tiny}c>{\tiny}c}
	\includegraphics[ width=\wb\linewidth, keepaspectratio,]{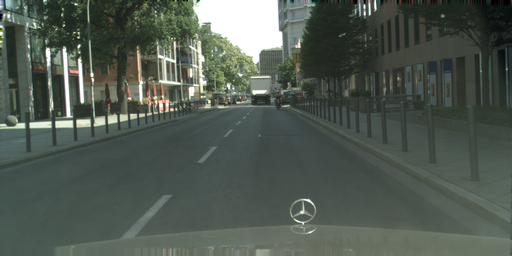}&%
	\includegraphics[ width=\wb\linewidth, keepaspectratio,]{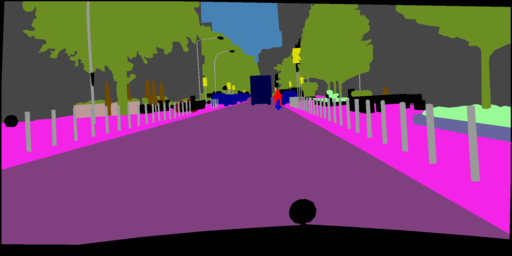}&%
	\includegraphics[ width=\wb\linewidth, keepaspectratio,]{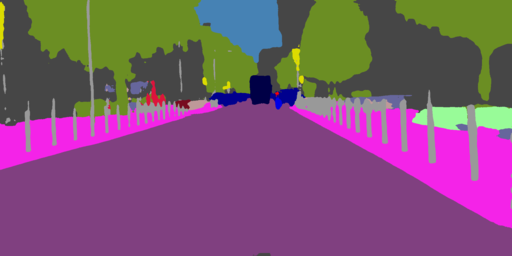}&%
	\includegraphics[ width=\wb\linewidth, keepaspectratio,]{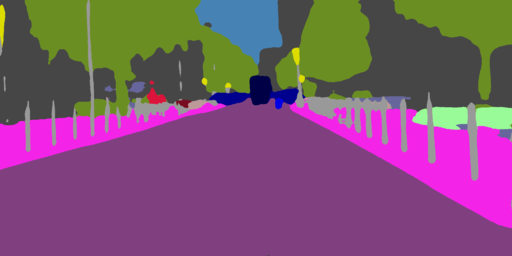}&%
	\includegraphics[ width=\wb\linewidth, keepaspectratio,]{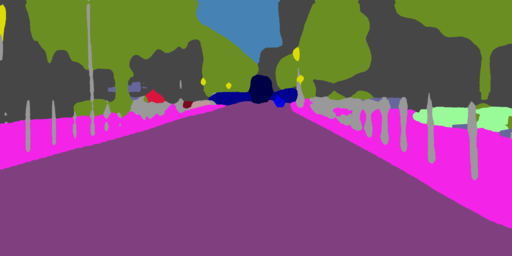}\\%
	\includegraphics[ width=\wb\linewidth, keepaspectratio,]{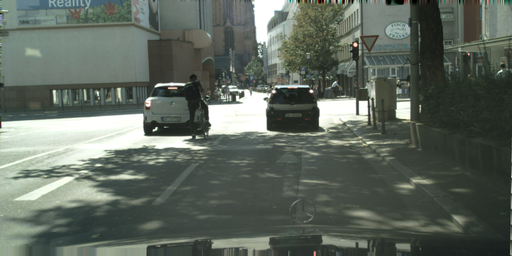}&%
	\includegraphics[ width=\wb\linewidth, keepaspectratio,]{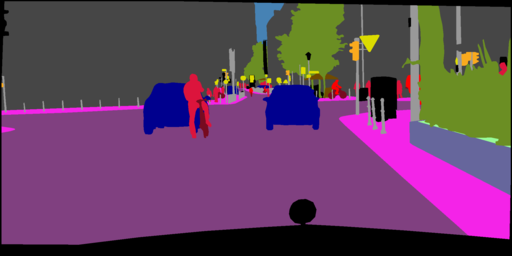}&%
	\includegraphics[ width=\wb\linewidth, keepaspectratio,]{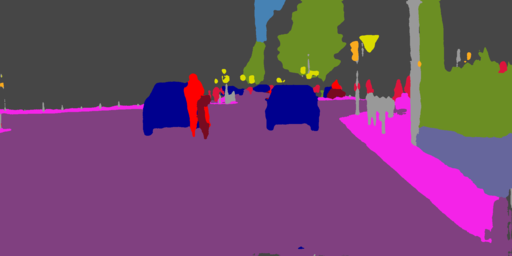}&%
	\includegraphics[ width=\wb\linewidth, keepaspectratio,]{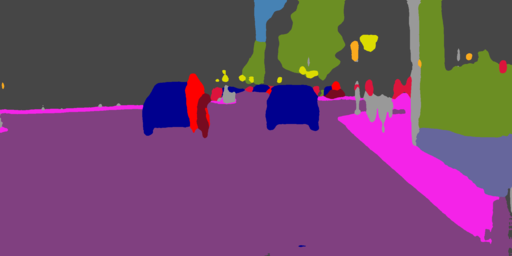}&%
	\includegraphics[ width=\wb\linewidth, keepaspectratio,]{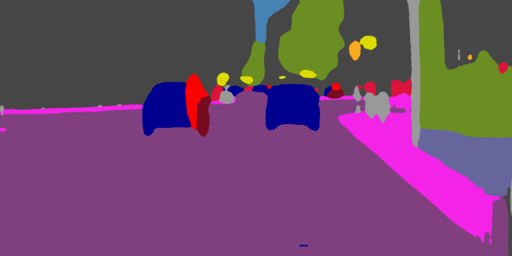}\\%
	\includegraphics[ width=\wb\linewidth, keepaspectratio,]{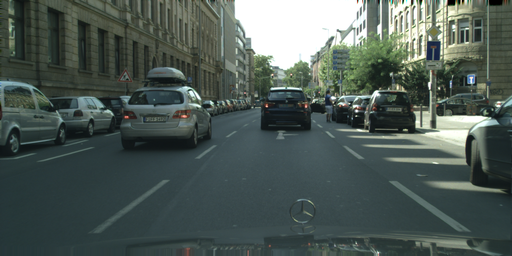}&%
	\includegraphics[ width=\wb\linewidth, keepaspectratio,]{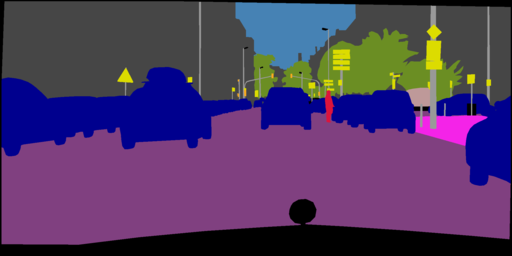}&%
	\includegraphics[ width=\wb\linewidth, keepaspectratio,]{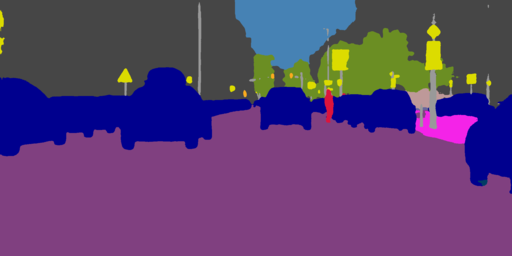}&%
	\includegraphics[ width=\wb\linewidth, keepaspectratio,]{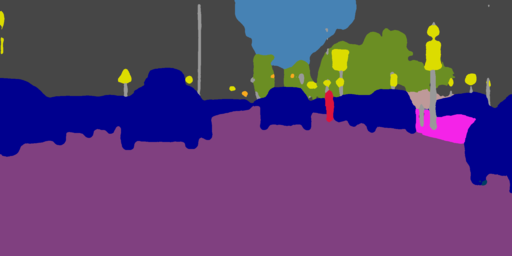}&%
	\includegraphics[ width=\wb\linewidth, keepaspectratio,]{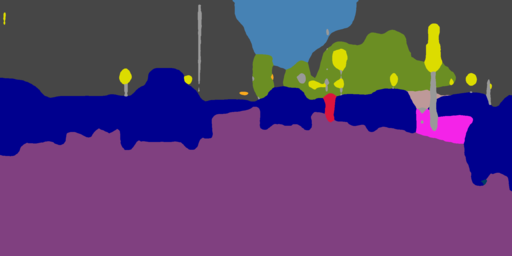}\\%
	\includegraphics[ width=\wb\linewidth, keepaspectratio,]{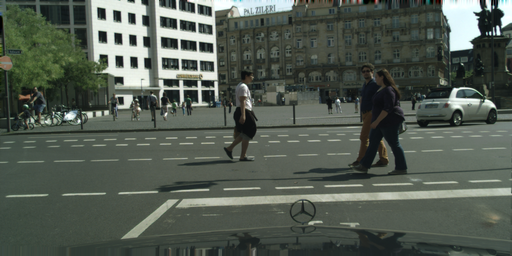}&%
	\includegraphics[ width=\wb\linewidth, keepaspectratio,]{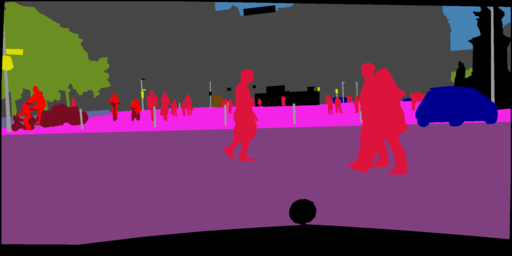}&%
	\includegraphics[ width=\wb\linewidth, keepaspectratio,]{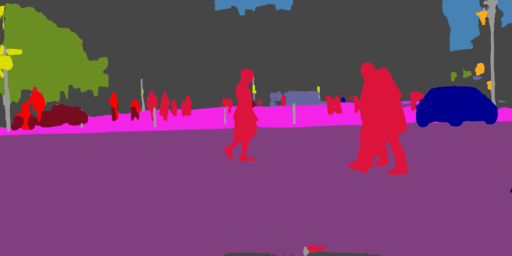}&%
	\includegraphics[ width=\wb\linewidth, keepaspectratio,]{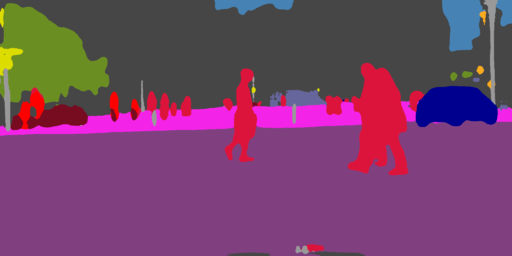}&%
	\includegraphics[ width=\wb\linewidth, keepaspectratio,]{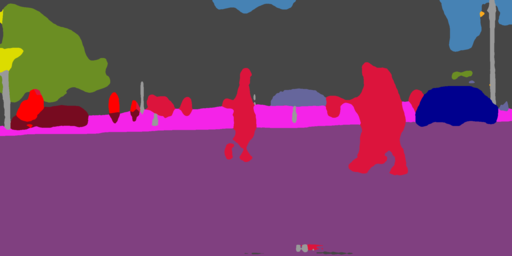}\\%
	\includegraphics[ width=\wb\linewidth, keepaspectratio,]{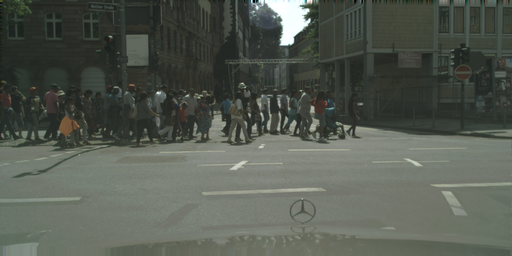}&%
	\includegraphics[ width=\wb\linewidth, keepaspectratio,]{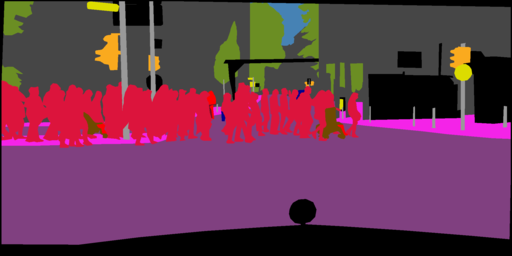}&%
	\includegraphics[ width=\wb\linewidth, keepaspectratio,]{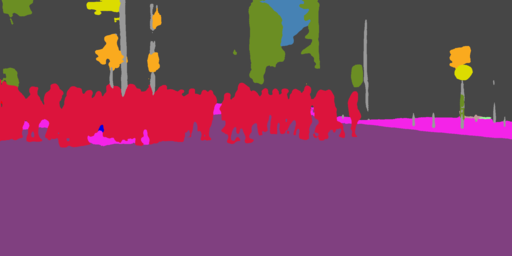}&%
	\includegraphics[ width=\wb\linewidth, keepaspectratio,]{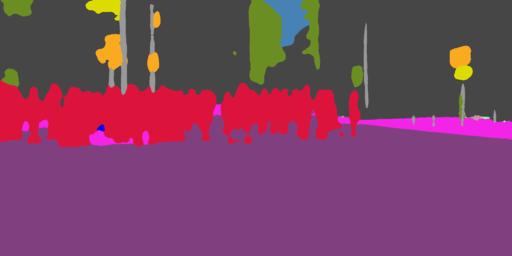}&%
	\includegraphics[ width=\wb\linewidth, keepaspectratio,]{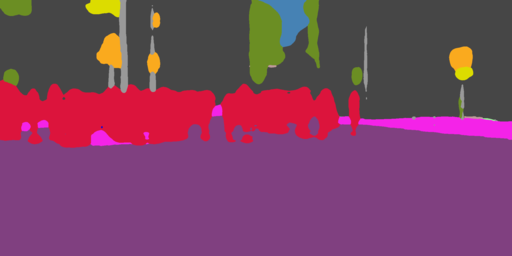}\\%
	\includegraphics[ width=\wb\linewidth, keepaspectratio,]{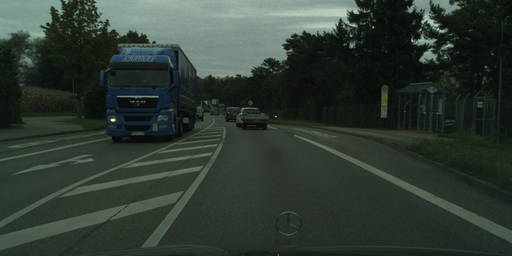}&%
	\includegraphics[ width=\wb\linewidth, keepaspectratio,]{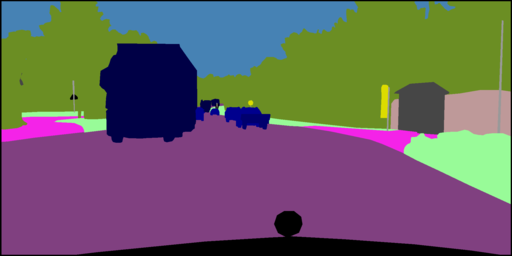}&%
	\includegraphics[ width=\wb\linewidth, keepaspectratio,]{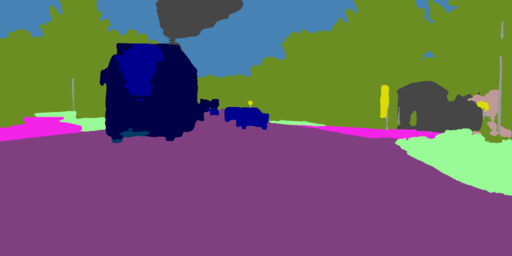}&%
	\includegraphics[ width=\wb\linewidth, keepaspectratio,]{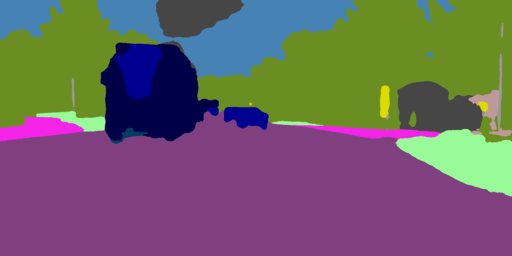}&%
	\includegraphics[ width=\wb\linewidth, keepaspectratio,]{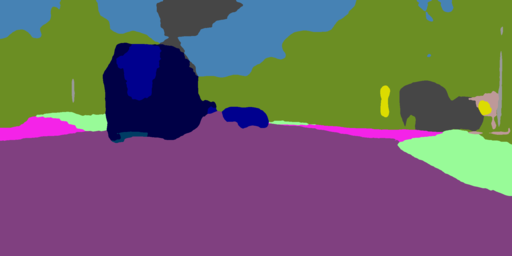}\\%
	&ground-truth&LRR-4x (8x) & LRR-4x (16x) & LRR-4x (32x)\\
\end{tabular}
\caption{Examples of semantic segmentation results on PASCAL VOC 2011 (top) and
Cityscapes (bottom) validation images. For each row, we show the input image, ground-truth and the
segmentation results of intermediate outputs of our LRR-4x model at the 32x, 16x and 8x layers.
For the PASCAL dataset we also show segmentation results of FCN-8s~\cite{long2015fcn}.}
\label{figure:qualitative}
\vspace{-0.15in}
\end{figure}

The images of Cityscapes are high resolution ($1024 \times 2048$) which makes
training challenging due to limited GPU memory.  We trained our model on a
random crops of size $1024 \times 512$.  At test time, we split each image to 2
overlapping windows and combined the predicted class probability maps.  We
did not use any CRF post-processing on this dataset.
Fig.~\ref{tab:cityscapes_test} shows evaluation of our model built on VGG-16
on the validation and test data. It achieves competitive performance on the test
data in comparison to the state-of-the-art methods, particularly on the category
level benchmark. Examples of semantic segmentation results on the validation images
are shown in Fig.~\ref{figure:qualitative}


\section{Discussion and Conclusions}

We have presented a system for semantic segmentation that utilizes two simple,
extensible ideas: (1) sub-pixel upsampling using a class-specific
reconstruction basis, (2) a multi-level Laplacian pyramid reconstruction
architecture that uses multiplicative gating to more efficiently blend
semantic-rich low-resolution feature map predictions with spatial detail from
high-resolution feature maps. The resulting model is simple to train and
achieves performance on PASCAL VOC 2012 test and Cityscapes that beats all but
two recent models that involve considerably more elaborate architectures based
on deep CRFs.  We expect the relative simplicity and extensibility of our
approach along with its strong performance will make it a ready candidate for
further development or direct integration into more elaborate inference
models.\\ 

\noindent{\bf Acknowledgements:} This work was supported by NSF grants
IIS-1253538 and DBI-1262547 and a hardware donation from NVIDIA.

\clearpage

\bibliographystyle{splncs03}
\bibliography{arxiv}

\begin{thebibliography}{10}
\providecommand{\url}[1]{\texttt{#1}}
\providecommand{\urlprefix}{URL }

\bibitem{BurtAdelson}
Burt, P.J., Adelson, E.H.: The laplacian pyramid as a compact image code. IEEE
  Transactions on Communications  31(4),  532--540 (1983)

\bibitem{carreiraijcv2012}
Carreira, J., Li, F., Sminchisescu, C.: Object recognition by sequential
  figure-ground ranking. IJCV  98(3),  243--262 (2012)

\bibitem{carreirasiminchisescupami2012}
Carreira, J., Sminchisescu, C.: {CPMC}: Automatic object segmentation using
  constrained parametric min-cuts. PAMI  34(7),  1312--1328 (2012)

\bibitem{chen2015edgenet}
Chen, L.C., Barron, J.T., Papandreou, G., Murphy, K., Yuille, A.L.: Semantic
  image segmentation with task-specific edge detection using cnns and a
  discriminatively trained domain transform. In: CVPR (2016)

\bibitem{chen2015deeplab}
Chen, L.C., Papandreou, G., Kokkinos, I., Murphy, K., Yuille, A.L.: Semantic
  image segmentation with deep convolutional nets and fully connected crfs. In:
  ICLR (2015)

\bibitem{chen2016deeplab}
Chen, L.C., Papandreou, G., Kokkinos, I., Murphy, K., Yuille, A.L.: Deeplab:
  Semantic image segmentation with deep convolutional nets, atrous convolution,
  and fully connected crfs. arXiv preprint arXiv:1606.00915  (2016)

\bibitem{chen2015attention}
Chen, L.C., Yang, Y., Wang, J., Xu, W., Yuille, A.L.: Attention to scale:
  Scale-aware semantic image segmentation. In: CVPR (2015)

\bibitem{Cordts2016Cityscapes}
Cordts, M., Omran, M., Ramos, S., Rehfeld, T., Enzweiler, M., Benenson, R.,
  Franke, U., Roth, S., Schiele, B.: The cityscapes dataset for semantic urban
  scene understanding. In: CVPR (2016)

\bibitem{Dai2014-eu}
Dai, J., He, K., Sun, J.: Convolutional feature masking for joint object and
  stuff segmentation. arXiv preprint arXiv:1412.1283  (2014)

\bibitem{dai2015boxsup}
Dai, J., He, K., Sun, J.: Boxsup: Exploiting bounding boxes to supervise
  convolutional networks for semantic segmentation. In: ICCV. pp. 1635--1643
  (2015)

\bibitem{denton2015deep}
Denton, E.L., Chintala, S., Fergus, R., et~al.: Deep generative image models
  using a laplacian pyramid of adversarial networks. In: NIPS. pp. 1486--1494
  (2015)

\bibitem{everingham2015pascal}
Everingham, M., Eslami, S.A., Van~Gool, L., Williams, C.K., Winn, J.,
  Zisserman, A.: The pascal visual object classes challenge: A retrospective.
  IJCV pp. 98--136 (2015)

\bibitem{gidaris2015object}
Gidaris, S., Komodakis, N.: Object detection via a multi-region and semantic
  segmentation-aware cnn model. In: ICCV. pp. 1134--1142 (2015)

\bibitem{hariharan2011semantic}
Hariharan, B., Arbel{\'a}ez, P., Bourdev, L., Maji, S., Malik, J.: Semantic
  contours from inverse detectors. In: ICCV. pp. 991--998 (2011)

\bibitem{hariharan2015hypercolumns}
Hariharan, B., Arbel{\'a}ez, P., Girshick, R., Malik, J.: Hypercolumns for
  object segmentation and fine-grained localization. In: CVPR. pp. 447--456
  (2015)

\bibitem{residualnetsarxiv2015}
He, K., Zhang, X., Ren, S., Sun, J.: Deep residual learning for image
  recognition. arXiv preprint arXiv:1512.03385  (2015)

\bibitem{he2016identity}
He, K., Zhang, X., Ren, S., Sun, J.: Identity mappings in deep residual
  networks. In: ECCV (2016)

\bibitem{kohli2009hierpotentials}
Kohli, P., Ladick´y, L., Torr, P.H.: Robust higher order potentials for
  enforcing label consistency. IJCV  82(3),  302--324 (2009)

\bibitem{koki2016supbound}
Kokkinos, I.: Pushing the boundaries of boundary detection using deep learning.
  In: ICLR (2016)

\bibitem{koltun2011efficient}
Kr\"ahenb\"uhl, P., Koltun, V.: Efficient inference in fully connected crfs
  with gaussian edge potentials. In: NIPS (2011)

\bibitem{lin2016piecewise}
Lin, G., Shen, C., Hengel, A.v.d., Reid, I.: Efficient piecewise training of
  deep structured models for semantic segmentation. In: CVPR (2016)

\bibitem{coco2014}
Lin, T.Y., Maire, M., Belongie, S., Hays, J., Perona, P., Ramanan, D.,
  Doll{\'a}r, P., Zitnick, C.L.: Microsoft coco: Common objects in context. In:
  ECCV. pp. 740--755 (2014)

\bibitem{liu2015parsing}
Liu, Z., Li, X., Luo, P., Loy, C.C., Tang, X.: Semantic image segmentation via
  deep parsing network. In: ICCV. pp. 1377--1385 (2015)

\bibitem{long2015fcn}
Long, J., Shelhamer, E., Darrell, T.: Fully convolutional networks for semantic
  segmentation. In: CVPR. pp. 3431--3440 (2015)

\bibitem{mostajabi2015zoomout}
Mostajabi, M., Yadollahpour, P., Shakhnarovich, G.: Feedforward semantic
  segmentation with zoom-out features. In: CVPR. pp. 3376--3385 (2015)

\bibitem{noh2015deconvnet}
Noh, H., Hong, S., Han, B.: Learning deconvolution network for semantic
  segmentation. In: ICCV. pp. 1520--1528 (2015)

\bibitem{piotrCVPR2016}
Pinheiro, P.O., Lin, T.Y., Collobert, R., Doll{\'a}r, P.: Learning to refine
  object segments. In: ECCV (2016)

\bibitem{textonboost}
Shotton, J., Winn, J., Rother, C., Criminisi, A.: Textonboost for image
  understanding: Multi-class object recognition and segmentation by jointly
  modeling texture, layout, and context. IJCV  81(1),  2--23 (2009)

\bibitem{simonyan2014very}
Simonyan, K., Zisserman, A.: Very deep convolutional networks for large-scale
  image recognition. arXiv preprint arXiv:1409.1556  (2014)

\bibitem{uhrig2016pixel}
Uhrig, J., Cordts, M., Franke, U., Brox, T.: Pixel-level encoding and depth
  layering for instance-level semantic labeling. arXiv preprint
  arXiv:1604.05096  (2016)

\bibitem{vedaldi15matconvnet}
Vedaldi, A., Lenc, K.: Matconvnet {--} convolutional neural networks for
  matlab. In: ICML (2015)

\bibitem{xie2015holistically}
Xie, S., Tu, Z.: Holistically-nested edge detection. In: ICCV. pp. 1395--1403
  (2015)

\bibitem{SongfanYangRamanan2015}
Yang, S., Ramanan, D.: Multi-scale recognition with dag-cnns. In: ICCV. pp.
  1215--1223 (2015)

\bibitem{Yu2015-yy}
Yu, F., Koltun, V.: Multi-scale context aggregation by dilated convolutions.
  arXiv preprint arXiv:1511.07122  (2015)

\bibitem{zeiler2014visualizing}
Zeiler, M.D., Fergus, R.: Visualizing and understanding convolutional networks.
  In: ECCV. pp. 818--833 (2014)

\bibitem{ZeilerICCV2011}
Zeiler, M.D., Taylor, G.W., Fergus, R.: Adaptive deconvolutional networks for
  mid and high level feature learning. In: ICCV. pp. 2018--2025 (2011)

\bibitem{zheng2015crfrnn}
Zheng, S., Jayasumana, S., Romera-Paredes, B., Vineet, V., Su, Z., Du, D.,
  Huang, C., Torr, P.H.: Conditional random fields as recurrent neural
  networks. In: ICCV. pp. 1529--1537 (2015)

\end{thebibliography}
\end{document}